\documentclass[11pt]{article}

\usepackage[letterpaper,margin=1in]{geometry}
\usepackage{times}
\usepackage[authoryear,round]{natbib}
\setcitestyle{citesep={;},aysep={,},yysep={;}}

\usepackage{amsmath,amsfonts,bm}

\def\eqref#1{equation~\ref{#1}}
\def\Eqref#1{Equation~\ref{#1}}

\def\1{\bm{1}}

\def\eps{{\epsilon}}

\DeclareMathAlphabet{\mathsfit}{\encodingdefault}{\sfdefault}{m}{sl}
\SetMathAlphabet{\mathsfit}{bold}{\encodingdefault}{\sfdefault}{bx}{n}

\newcommand{\E}{\mathbb{E}}

\newcommand{\R}{\mathbb{R}}

\DeclareMathOperator*{\argmin}{arg\,min}

\usepackage{algorithm}
\usepackage[noend]{algpseudocode}
\algrenewcommand\algorithmicrequire{\textbf{Input:}}
\algrenewcommand\algorithmicensure{\textbf{Output:}}
\newcommand{\algrule}[1][.2pt]{\par\vskip.2\baselineskip\hrule height #1\par\vskip.2\baselineskip}
\usepackage[colorlinks=true,linkcolor=blue,citecolor=blue,urlcolor=blue]{hyperref}
\usepackage{url}
\usepackage{graphicx}
\graphicspath{{figures/}}
  \usepackage{nth}
  \usepackage{intcalc}

  \newcommand{\cNIPS}[1]{Advances\ in\ Neural\ Information\ Processing\ Systems\ (NeurIPS)}

  \newcommand{\cAAAI}[1]{AAAI\ Conference\ on\ Artificial\ Intelligence\ (AAAI)}

\usepackage{booktabs}
\usepackage{longtable}
\usepackage{amsmath}
\usepackage{amssymb}
\usepackage{amsthm}
\usepackage{cleveref}

\DeclareMathOperator{\SW}{SW}
\newcommand{\Unif}{\mathrm{Unif}}
\newcommand{\Tbar}{\bar{T}}
\newcommand{\qhat}{\hat{q}}

\newcommand{\defeq}{:=}
\newcommand{\norm}[1]{\left\lVert #1 \right\rVert}

\newcommand{\paren}[1]{\left(#1\right)}

\newtheorem{theorem}{Theorem}
\newtheorem{corollary}{Corollary}
\newtheorem{lemma}[theorem]{Lemma}
\newtheorem{proposition}[theorem]{Proposition}

\newtheorem*{reptheoreminner}{Theorem \repnumber}
\newtheorem*{replemmainner}{Lemma \repnumber}
\newenvironment{reptheorem}[2][]{%
  \def\repnumber{\ref{#2}}%
  \if\relax\detokenize{#1}\relax
    \begin{reptheoreminner}%
  \else
    \begin{reptheoreminner}[#1]%
  \fi
}{\end{reptheoreminner}}
\newenvironment{replemma}[2][]{%
  \def\repnumber{\ref{#2}}%
  \if\relax\detokenize{#1}\relax
    \begin{replemmainner}%
  \else
    \begin{replemmainner}[#1]%
  \fi
}{\end{replemmainner}}

\newcommand{\methodname}{\texttt{SPLICE}}

\title{Differentially Private Synthetic Data from Sliced Chebyshev Moments}

\author{%
  Lucas Rosenblatt\\
  Department of Computer Science\\
  Williams College\\
  \texttt{lr21@williams.edu}
  \and
  R. Teal Witter\\
  Department of Mathematical Sciences\\
  Claremont McKenna College\\
  \texttt{rtealwitter@cmc.edu}
  \and
  Christopher Musco\\
  Department of Computer Science and Engineering\\
  New York University\\
  \texttt{cmusco@nyu.edu}
}
\date{}

\begin{document}

\maketitle

\begin{abstract}
Research on practical differentially private (DP) synthetic data has largely focused on methods that approximately preserve low-dimensional marginals of a target distribution, and can thus only hope to preserve higher-dimensional, global structure as a lucky by-product. At the same time, existing methods that target global accuracy guarantees pay a price exponential in the data dimension, $d$.

In this work, we address this ``curse of dimensionality'' by presenting a new DP synthetic data algorithm, \methodname{}, that comes with a compelling global distributional guarantee, even in high-dimensions, and moreover, is practically state-of-the-art. Our algorithm is based on measuring noisy \textit{sliced} Chebyshev polynomial moments of the target distribution and fitting a distribution to match, a twist on recent polynomial methods for 1-D synthetic data.

We prove that, given a dataset of $n$ examples, \methodname{} returns a private distribution with Sliced Wasserstein-1 distance $\tilde{O}(1/\sqrt{n})$ from the target, for constant privacy parameters. Notably, this bound is independent of the data dimension, $d$.
Along the way, we prove a moment matching result of independent interest: for a distribution on the unit ball, matching the first $O(1/\alpha)$ Chebyshev moments along $\tilde{O}(d/\alpha^2)$ random directions guarantees $\alpha$ Sliced Wasserstein error.

Finally, we provide a practical implementation of the algorithm that borrows tools from the tensor decomposition literature and achieves state-of-the-art performance compared to prior DP synthetic data methods on an extensive benchmark.
\end{abstract}

\section{Introduction}
\label{sec:intro}
The generation of high-utility differentially private (DP) synthetic data has become a focal point of research into DP data release \citep{cummings2024advancing}. In the basic setup, we are given a dataset $\mathcal{X} = \{x_1, \ldots, x_n\} \subset \R^{d}$ and wish to privately compute a distribution $q$ that well-approximates the empirical distribution of $\mathcal{X}$, which we denote by $p$ (i.e., $p$ is uniform over $x_1, \ldots, x_n$). The hope is that $q$, or samples from $q$, can be used as a drop-in replacement for $\mathcal{X}$ in downstream machine learning and data analysis tasks, without the need for any further differential privacy mechanisms.

To guarantee strong performance on downstream tasks, a natural goal is to develop algorithms that ensure $q$ is close to $p$ with respect to some meaningful statistical distance, like Wasserstein distance or some other integral probability metric. There has been significant theoretical work targeting such guarantees \citep{He:2023, Boedihardjo:2024, feldman2024instance, musco2025sharper, DingHeZhu:2026}. However, all of this work obtains bounds that suffer exponentially in the data dimension, $d$, so it has yielded few practically competitive algorithms.

Indeed, empirically oriented synthetic data methods give up on strong distribution guarantees, focusing on achieving practical utility on realistic downstream tasks \citep{PrivBayes:2017,liu2021iterative, mckenna2022aim, liu2023generating}. Many of the best practical methods are based on noisily measuring only low-order marginals of $p$ (involving at most $2$ or $3$ variables \citep{mckenna2022aim}) and returning a synthetic distribution that matches these marginals (the ``Select-Measure-Estimate'' paradigm \citep{ponomareva2025dp}). While they sometimes come with workload-specific error guarantees, it is not surprising that higher-dimensional correlations and structure are missed.

\begin{figure}[t]
  \centering
  \includegraphics[width=0.4\linewidth]{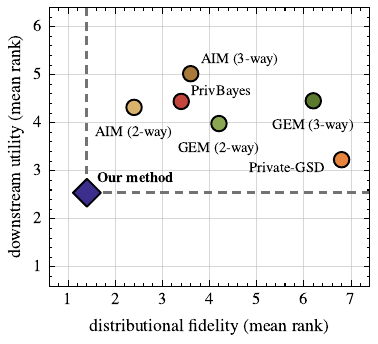}
  \caption{This plot summarizes a comparison of state-of-the-art DP synthetic data algorithms. Each method is evaluated on the 48 downstream train-on-synthetic, test-on-real tasks and ranked according to accuracy. The distribution returned by each method was also ranked using five fidelity metrics (1-way, 2-way, and 3-way marginal TVD, kernel MMD, and sliced $W_1$). The average rank determines the $x$ and $y$ coordinates. Lower is better on both axes.}
  \label{fig:overall}
\end{figure}

\subsection{Our Contributions}
This work seeks to address this ``curse of dimensionality'' in synthetic data generation. In short, we introduce a new algorithm, \methodname{}, that enjoys a strong distributional error guarantee, even in high dimensions, and moreover, offers state-of-the-art practical performance on downstream tasks.

\methodname{} is inspired by a recent approach for low-dimensional synthetic data generation based on matching polynomial moments (specifically, noisy Chebyshev polynomial moments) \citep{WangEtAl:2016}.
This approach yields near-optimal bounds in terms of Wasserstein-1 ($W_1$) distance: \cite{musco2025sharper} show that, for a dataset with $n$ 1-d points, moment matching yields an $(\eps,\delta)$-DP distribution $q$ such that $W_1(p,q) \leq \tilde{O}(\sqrt{\log(1/\delta)}/\epsilon n)$. At the same time, Chebyshev moment matching falls under the ``Select-Measure-Estimate'' paradigm, so is easy to implement efficiently.

Unfortunately, however, like other approaches listed above, Chebyshev moment matching does not readily scale to high dimensions. For $d$-dimensional data, the bound of \cite{musco2025sharper} degrades to $\tilde{O}(1/n^{1/d})$ for constant $\epsilon,\delta$. This is vacuous for $d \gtrsim \log n$ and weak even for small constant $d$. Moreover, the bound cannot be improved: \cite{Boedihardjo:2024} show that no method can privately achieve Wasserstein-1 error below ${O}(1/n^{1/d})$.

\paragraph{Algorithmic Contribution.}
The key issue with prior work is that obtaining small Wasserstein-1 error requires matching multivariate products of Chebyshev polynomials, the number of which grows exponentially in the number of data variables.
In \methodname{}, we propose to instead measure \emph{sliced} Chebyshev polynomial moments, which for any unit vector $\theta\in \R^d$, take the form:
\begin{align*}
\E_{x \sim \Unif(\mathcal{X})}\left[T_j(\theta^T x)\right],
\end{align*}
where $T_j$ is a Chebyshev polynomial of degree $j$. The method computes moments of this form for $L$ different random vectors $\theta_1, \ldots, \theta_L$, where $L$ scales polynomially in $d$. It then adds noise to those moments to ensure differential privacy, and finally fits a distribution $q$ that matches the noisy moments. In other words, the method again fits into the popular ``Select-Measure-Estimate'' paradigm, but replaces low-degree marginal measurements with \emph{sliced} moment measurements.

Our main theoretical result is to prove that, while Wasserstein distance is out of reach, \methodname{} yields a strong bound on the \emph{sliced Wasserstein-1 distance} between our private distribution $q$ and target $p$. Letting $W_1$ denote the one-dimensional Wasserstein-1 distance and $\mathcal{S}^{d-1}$ denotes the unit sphere in $d$ dimensional Euclidean space, sliced Wasserstein-1 distance is defined as:
\begin{align}
   \label{def:sw}
   \SW_1(p, q) \defeq \E_{\theta \sim \Unif(\mathcal{S}^{d-1})}\left[ W_1(\bar{p}_{\theta}, \bar{q}_{\theta}) \right],
\end{align}
where $\bar{p}_{\theta}$ (analogously, $\bar{q}_{\theta}$) is the pushforward distribution obtained by projecting $p$ onto $\theta \in \mathcal{S}^{d-1}$:
\begin{align}
\label{eq:pushforward}
\bar{p}_{\theta}(A) = p(\{x : \theta^\top x \in A\}) \text{ for any measurable set $A \subseteq \R$.}
\end{align}
In other words, $\SW_1(p, q)$ is the average Wasserstein-1 distance between one-dimensional projections of $p$ and $q$.
Originally introduced as a tractable approximation for the standard Wasserstein-1 distance \citep{rabin2011wasserstein, bonneel2015sliced}, sliced Wasserstein distance has been widely adopted across machine learning, especially in the generative model literature, as a valuable measure in its own right \citep{deshpande2018generative,deshpande2019max,kolouri2018sliced,kolouri2019generalized,liutkus2019sliced,nadjahi2020statistical,sebag2025differentially,gu2025private}.

Concretely, we prove the following bound, which is stated more formally in \Cref{sec:splice} and proven in \Cref{app:dp_utility}:

\begin{reptheorem}{thm:dp}
Let $\mathcal{B}^d$ denote the unit ball in $\R^d$. There is an $(\epsilon,\delta)$-differentially private algorithm (\methodname{}) that, given a dataset $\mathcal{X} = \{x_1, \ldots, x_n\} \subset \mathcal{B}^d$ with $p = \Unif(\mathcal{X})$, returns $q$ such that
\begin{align*}
\tag{\ref{eq:dp_dimension_free_rate}}
    \SW_1(p,q)
    &\leq \tilde{O}\left(\frac{1}{\sqrt{n\epsilon}}\right), & &\text{with probability at least } 99/100.
\end{align*}
\end{reptheorem}
Critically, the bound in \Cref{thm:dp} is independent of the data dimension $d$: sliced Wasserstein error $\alpha$ requires only $n = \tilde{O}(1/(\epsilon\alpha^2))$ data examples, though we allow the number of sampled directions to still depend on $d$. In contrast, as discussed above, obtaining even constant Wasserstein-1 error requires $n = 2^{\Omega(d)}$ examples \citep{Boedihardjo:2024}. The lack of dimension dependence in \Cref{thm:dp} mirrors existing dimension-free bounds on the convergence of an empirical measure to the true distribution under sliced Wasserstein distance \citep{nadjahi2020statistical,manole2022minimax,xi2022distributional,boedihardjo2024sharp}.

Key to the proof of \Cref{thm:dp} is a moment matching result of potentially independent interest. We show that, if two high-dimensional distributions, $p$ and $q$, match on their first $1/\alpha$ moments along $L = \tilde{O}(d/\alpha^2)$ random slices, then $\SW_1(p,q) \leq O(\alpha)$. This generalizes the well-known fact that, in one dimension, matching the first $1/\alpha$ moments of two distributions guarantees $W_1(p,q) \leq \alpha$ \citep{KongValiant:2017,BravermanKrishnanMusco:2022}. The key technical challenge in establishing this result is to show that a polynomial bound on the required number of slices, $L$, suffices. A naive analysis suggests that $L$ needs to scale exponentially in $d$ to cover the $d$ dimensional sphere.

\paragraph{Empirical Contribution.}
\Cref{thm:dp} tells us how many data examples, $n$ suffice to build an accurate synthetic distribution, $q$, but does not say anything about the complexity of representing or learning that distribution from noisy moment measurements. To make our results practical, we choose to represent $q$ as a mixture of $K$ product distributions: in each component of the mixture, discrete random variables follow independent distributions, and continuous random variables are drawn from a single joint Gaussian distribution with low-rank-plus-diagonal covariance. Dependencies between discrete variables are introduced by mixing multiple such distributions.

The class of distributions we consider thus involves $O(d)$ parameters per component, so $O(Kd)$ in total. We fit these parameters using gradient descent, relying on the key fact that the distribution's sliced Chebyshev moments have an exact, differentiable closed form.
Empirically, using datasets from the extensive TALENT tabular data benchmark \citep{JMLR:v26:25-0512}, we evaluate \methodname{} with this chosen representation for $q$. Our empirical results are quite striking.

Concretely, we find that the mechanism gives the best regression utility of any method among a large set of state-of-the-art benchmarks, at every privacy budget. It also offers the best error on 2-way and 3-way marginal fidelity for $\epsilon \geq 1$, despite not targeting marginals directly, and achieves the best causal structure recovery by a significant margin. These results are summarized in \Cref{fig:overall} at $\eps = 1$, where \methodname{} outperforms strong marginal-based baselines on both composite axes simultaneously, distributional fidelity and downstream utility, with detailed experiments contained in \Cref{sec:experiments}. Finally, we note that the runtime of \methodname{} is competitive with the fastest baselines and dramatically faster than the slowest for typical dataset sizes.

\paragraph{Roadmap.} \Cref{sec:moment_matching} provides our key technical result, that sliced moments along a small number of random directions suffice to control $\SW_1$. \Cref{sec:splice} leverages this result to obtain the DP guarantee of \Cref{thm:dp}. \Cref{sec:tractable} instantiates the algorithm with a practical learned representation, and \Cref{sec:experiments} empirically tests it. \Cref{app:related} surveys related work in additional depth.

\section{Sliced Chebyshev Moment Matching}
\label{sec:moment_matching}
We begin with basic preliminaries on the Wasserstein distance and moment matching. Throughout, our goal will be to approximate a distribution $p$, which in the case of differentially private synthetic data, will be taken to be the uniform distribution over some dataset $\mathcal{X} = \{x_1,\ldots, x_n\} \subset \R^{d}$.

Let $q$ be some other distribution. The Wasserstein-1 distance between $p$ and $q$ is defined as follows:
\begin{align*}
W_1(p,q) \defeq \inf_{z\in Z(p,q)} \left[\E_{(x,y)\sim z} \|x-y\|_2\right],
\end{align*}
where $\|\cdot \|_2$ is the Euclidean distance and $Z(p,q)$ is the set of all couplings between $p$ and $q$.

A powerful fact is that, if two 1-dimensional distributions ${p}$ and ${q}$ supported\footnote{The support assumption is to simplify notation, since we can always rescale the data to lie in $[-1,1]$. Without scaling, the error would be $O(1/k)$ times the diameter of the support. For high-dimensional distributions, we will likewise assume that $p$ and $q$ are supported on the unit ball $\mathcal{B}^d$ to reduce notational complexity.} on $[-1,1]$ match on their first $k$ polynomial moments (i.e., $\E_{x \sim {p}}[x^j] = \E_{x \sim {q}}[x^j]$ for all $j \in 1, \ldots, k$), then $W_1({p}, {q}) \leq O(1/k)$ \citep{KongValiant:2017,BravermanKrishnanMusco:2022}. This claim follows from combining Jackson's theorem \citep{Jackson:1912}, which says that any $1$-Lipschitz function on $[-1,1]$ can be approximated to error $O(1/k)$ by a degree $k$ polynomial, with the Kantorovich-Rubinstein dual view of Wasserstein distance, which states that $W_1({p}, {q}) = \max_{1 \text{ Lipschitz } f} \E_{x\sim {p}}[f(x)] - \E_{x\sim {q}}[f(x)]$.

Importantly, this moment matching claim can be made robust when using a basis of polynomial moments besides the standard basis (which is ill-conditioned). For example, let $T_j$ denote the $j^\text{th}$ Chebyshev polynomial of the first kind:
\begin{align*}
T_0(x) &= 1, & T_1(x) &= x, & T_{j+1}(x) &= 2 x \cdot T_j(x) - T_{j-1}(x) \text{ for } j \geq 1.
\end{align*}
Let $m_j(p) = \E_{x\sim {p}} \left[T_j(x)\right]$ denote the $j^\text{th}$ Chebyshev moment.
A recent result shows the following.
\begin{theorem}[{\cite[Theorem 1]{musco2025sharper}}]
\label{thm:mmrs1d}
Let $p, q$ be probability measures on $[-1, 1]$. For any positive integer $k$,
\begin{align}
   \label{eq:gamma1d}
   W_1(p, q) &\leq \frac{36}{k} + \Gamma_k(p,q), & &\text{where} &\Gamma_k(p,q) &= \left(\sum_{j=1}^k \frac{1}{j^2} \left(m_j(p) - m_j(q)\right)^2\right)^{1/2}.
\end{align}
\end{theorem}
\Cref{thm:mmrs1d} is useful when we wish to recover a distribution $p$ from \emph{noisy} approximations to its moments. In particular, suppose we know the first $k$ Chebyshev moments up to cumulative error $\Delta_k$ (in the sense of \eqref{eq:gamma1d}) and find $q$ matching them to error $\Delta_k$ (such a $q$ always exists, since $p$ is a valid choice). Because $\Gamma_k$ is a norm satisfying triangle inequality, any such $q$ satisfies $W_1(p,q) \leq 36/k + 2\Delta_k$.
This argument was used by \cite{musco2025sharper} to obtain a near-optimal algorithm for 1-dimensional DP synthetic data based on noising Chebyshev moments with the Gaussian mechanism \citep{DworkRoth:2014}. They show that, for a dataset with $n$ 1-dimensional points, it is possible to return an $(\epsilon,\delta)$-differentially private distribution $q$ satisfying $W_1(p,q) \leq \tilde{O}(\sqrt{\log(1/\delta)}/\epsilon n)$.

\subsection{Sliced Moment Matching}
As discussed, extending \Cref{thm:mmrs1d} to $d$ dimensions requires matching a number of moments exponential in $d$, resulting in Wasserstein error scaling as $O(1/n^{1/d})$ \citep{musco2025sharper}. Our goal is to avoid this poor dependence on $d$ by focusing on the relaxed sliced Wasserstein-1 distance defined in \eqref{def:sw}, noting that we always have $\SW_1(p,q) \leq W_1(p,q)$ \citep{carlier2025sharp}.

Towards this goal, we can immediately obtain the following corollary of
\Cref{thm:mmrs1d} for high dimensional distributions. In particular, let $p,q$
be probability measures on the unit ball $\mathcal{B}^d=\{x\in\R^d:\|x\|_2\leq1\}$
and define $m_j(\theta,p)=\E_{x\sim p}[T_j(x^\top\theta)]$ to be the
$j^\text{th}$ Chebyshev moment of the one-dimensional projection of $p$
as defined in \eqref{eq:pushforward}. Define $m_j(\theta,q)$ analogously.
For any probability measure $q$ on $\mathcal{B}^d$, define its weighted
moment error in direction $\theta$ by
\begin{align}
\label{eq:direct_slice_error}
    \Gamma_k(\theta,p,q)
    \defeq
    \left(
        \sum_{j=1}^k \frac{1}{j^2}
        \paren{m_j(\theta,p)-m_j(\theta,q)}^2
    \right)^{1/2}.
\end{align}

Applying \Cref{thm:mmrs1d} to the one-dimensional projections and averaging
over directions gives:
\begin{corollary}
\label{cor:sliced_mmrs}
Let $p,q$ be probability measures on $\mathcal{B}^d$. For any positive integer $k$,
\begin{align}
\label{eq:sliced_moment_bound}
    \SW_1(p,q)&\leq\frac{36}{k}+\bar\Gamma_k(p,q),
    & &\text{where}
    &\bar\Gamma_k(p,q)&=\E_{\theta\sim\Unif(S^{d-1})}[\Gamma_k(\theta,p,q)].
\end{align}
\end{corollary}
\Cref{cor:sliced_mmrs} is not immediately algorithmically useful, as it
requires us to control (on average) how closely $q$ matches the first $k$
Chebyshev moments of $p$ in every direction $\theta\in\mathcal{S}^{d-1}$.
Our main technical tool is a proof that it suffices to instead only match moments for a small random
sample of directions $\theta_1,\ldots,\theta_L$ drawn uniformly from
$\mathcal{S}^{d-1}$. In particular, we show:
\begin{theorem}
\label{thm:average_control}
Let $p$ be a probability measure on $\mathcal{B}^d$, let $L$ be a positive
integer, let $\beta\in(0,1)$, and let
$\theta_1,\ldots,\theta_L$ be independent draws from
$\Unif(S^{d-1})$. For any positive integer $k$, with probability at least
$1-\beta$ over the choice of
$\theta_1,\ldots,\theta_L$, simultaneously for all probability measures $q$
on $\mathcal{B}^d$,
\begin{equation}
\label{eq:direct_main_bound}
\begin{aligned}
    \SW_1(p,q)
    &\leq \frac{36}{k}+\tilde{\Gamma}_{k,L}(p,q)
        +\frac{32}{\beta}\sqrt{\frac{d\log(6Lk^2)}{L}},
    &\text{where~~}
    &\tilde{\Gamma}_{k,L}(p,q)
    =\frac{1}{L}\sum_{\ell=1}^L\Gamma_k(\theta_\ell,p,q).
\end{aligned}
\end{equation}
\end{theorem}
A consequence of this theorem is that, if we compute the first $k$ moments of $p$ along
$L=\tilde O(dk^2)$ random directions, the resulting information suffices
to find a distribution that is $O(1/k)$ close to $p$ in sliced Wasserstein
distance. In particular, we return any distribution that matches all
of the resulting $kL$ moments (so $\tilde\Gamma_{k,L}(p,q)=0$). Plugging into
\Cref{thm:average_control}, we obtain $\SW_1(p,q)=O(1/k)$.

However, the theorem is even more flexible in that, like \Cref{thm:mmrs1d}, it allows
us to return a $q$ that only approximately matches $p$'s sliced moments.
Approximation will be necessary in our application to differentially private data generation in \Cref{sec:splice}, since we only have access to noised moments.

Theorem \ref{thm:average_control} is proven by establishing that
$\bar\Gamma_k(p,q) = \E_{\theta\sim\Unif(S^{d-1})}\left[\Gamma_k(\theta,p,q)\right]$ is well approximated by
$\tilde\Gamma_{k,L}(p,q) = \frac{1}{L}\sum_{\ell=1}^L \Gamma_k(\theta_\ell,p,q)$ for all $q$. In particular, we show:
\begin{lemma}
\label{lem:error_matching}
Fix any positive integer $k$ and any probability measure $p$ on
$\mathcal{B}^d$. Let $L$ be a positive integer, let $\beta\in(0,1)$, and let
$\theta_1,\ldots,\theta_L$ be independent draws from
$\Unif(S^{d-1})$. With probability at least $1-\beta$ over the choice of
$\theta_1,\ldots,\theta_L$, simultaneously for all probability measures $q$
on $\mathcal{B}^d$,
\begin{align}
\label{eq:direct_uniform_approximation}
\left|
    \frac{1}{L}\sum_{\ell=1}^L \Gamma_k(\theta_\ell,p,q)
    - \E_{\theta\sim\Unif(S^{d-1})}\left[\Gamma_k(\theta,p,q)\right]
\right| \leq
\frac{32}{\beta}\sqrt{
    \frac{d\log(6Lk^2)}{L}
}.
\end{align}
\end{lemma}

\begin{proof}[Proof of \Cref{thm:average_control}]
By \Cref{cor:sliced_mmrs},
$\SW_1(p,q)\leq36/k+\bar\Gamma_k(p,q)$. By
Lemma~\ref{lem:error_matching},
$\bar\Gamma_k(p,q) \leq\tilde\Gamma_{k,L}(p,q) +\frac{32}{\beta}\sqrt{\frac{d\log(6Lk^2)}{L}}$ for all $q$, which immediately gives the theorem.
\end{proof}
\Cref{lem:error_matching} is proven in \Cref{app:moment_matching}. A natural proof approach would be to apply a concentration bound for each fixed distribution $q$, then extend to all distributions using an $\epsilon$-net argument. However, directly discretizing the space of distributions introduces an exponential dependence on $d$, since representing a distribution on the $d$-dimensional ball requires roughly $2^{O(d)}$ parameters. We avoid this issue by using symmetrization and Gaussian comparison arguments to reduce the problem to bounding a random linear combination of empirical moments. Since moments are linear in $q$, the resulting expression is maximized by a distribution that places all its mass at a single point. We can thus union bound over the $d$-dimensional ball, which has log-covering number just $O(d)$, leading to a linear dependence in the required number of direction samples, $L$.

\section{The \methodname{} Algorithm for DP Synthetic Data}
\label{sec:splice}
\Cref{thm:average_control} suggests a route to differentially private synthetic data that mirrors the one-dimensional mechanism of \citet{musco2025sharper}: compute sliced moments of the target dataset, intentionally add noise to those moments to ensure privacy, and construct a distribution $q$ that matches the noised moments. We present this approach formally in \Cref{alg:dp} below, showing how to calibrate the noise, and we then analyze the resulting accuracy of the method for any given privacy parameters.

\begin{algorithm}[ht]
\caption{Private Sliced Chebyshev Moment Matching}
\label{alg:dp}
\begin{algorithmic}[1]
\Require Dataset $\mathcal{X} = (x_1, \ldots, x_n) \in (\mathcal{B}^d)^n$; privacy parameters $\epsilon, \delta\in(0,1)$; integers $k, L \geq 1$.
\Ensure $(\epsilon,\delta)$-differentially private distribution $q$ approximating $p = \Unif(x_1, \ldots, x_n)$.
\algrule
\State \textbf{Sample directions.} Sample $\theta_1, \ldots, \theta_L \overset{\text{i.i.d.}}{\sim} \Unif(S^{d-1})$.
\State \textbf{Compute noisy moments.} For all $\ell \in 1, \ldots, L$ and $j \in 1, \ldots, k$, compute $\tilde{m}_{j,\ell} = \E_{x\sim p}[T_j(x^\top\theta_{\ell})] + \eta_{j,\ell}$, where $\eta_{j,\ell} \sim \mathcal{N}(0, j\sigma^2)$ with $\sigma^2 = {8L(1+\log k)\log(1.25/\delta)}/{(n^2\epsilon^2)}$.
\State \textbf{Match.} Using the sampled directions and noisy moments, return a distribution
\begin{align*}
    q\in\argmin_{r\text{ a probability measure on }\mathcal B^d}
        \frac1L\sum_{\ell=1}^L\sum_{j=1}^k\frac{(\tilde m_{j,\ell}-m_j(\theta_\ell,r))^2}{j^2}.
\end{align*}
\end{algorithmic}
\end{algorithm}

We remark that, as in \cite{musco2025sharper}, more noise is added to higher-degree polynomial moments, since these moments play a diminishing role in capturing global structure in $p$. Additionally, we highlight that we intentionally do not specify the details of the final ``Match'' step of the algorithm. Our goal in this section is to show that, if we solve the minimization problem shown, then $q$ will be close to $p$ in sliced Wasserstein distance with high probability. This is our main utility result. In the subsequent section, we describe an efficient heuristic for solving the problem in practice.

We first establish the privacy of \Cref{alg:dp}. Doing so is straight-forward, as $q$ only depends on $p$ through the noisy moments, $\tilde m_{j,\ell}$. These moments, in turn, are obtained by adding $\mathcal{N}(0,\sigma^2)$ noise (i.e., applying the standard Gaussian mechanism) to the vector valued function:
\begin{align}
\label{eq:dp_query_vector}
f(\mathcal{X})_{(\ell, j)} = \frac{1}{\sqrt{j}}\cdot\frac{1}{n}\sum_{i=1}^n T_j(\theta_\ell^\top x_i) \qquad \text{for } \ell \in 1, \ldots, L \text{ and } j \in 1, \ldots, k.
\end{align}
Note the scaling by $1/\sqrt{j}$: this ensures that the addition of $\mathcal{N}(0,j\sigma^2)$ noise in \Cref{alg:dp} is equivalent to adding $\mathcal{N}(0,\sigma^2)$ noise to each coordinate of $f(\mathcal{X})$.

Proving privacy thus amounts to bounding the $\ell_2$ sensitivity of $f(\mathcal{X})$. We do so under the standard replacement adjacency model, meaning that neighboring datasets differ in one record.
Using that $|T_j(z)| \leq 1$ for all $z \in [-1,1]$, we can bound the $\ell_2$ sensitivity of this function as:
\begin{align*}
\Delta_{2,f}^2 = \max_{\text{neighboring datasets } X, X'} \|f(X) - f(X')\|_2^2 \leq \sum_{\ell=1}^L \sum_{j=1}^k \frac{4}{jn^2} \leq \frac{4L}{n^2}(1 + \log k),
\end{align*}
where we used $\sum_{j=1}^k1/j\leq1+\log k$.
\citet[Theorem 3.22]{DworkRoth:2014} show that $(\epsilon,\delta)$-differential privacy can be achieved by adding Gaussian noise with variance $\Delta_{2,f}^2\cdot {2\log(1.25/\delta)}/{\epsilon^2}$, which equals $\sigma^2$. We thus conclude:
\begin{proposition}[Privacy]
\label{prop:privacy}
For $\epsilon,\delta \in (0,1)$, \Cref{alg:dp} is $(\epsilon,\delta)$-differentially private.
\end{proposition}
Next, we leverage the results of the previous section to show that \Cref{alg:dp} also achieves a strong utility guarantee. I.e., that $q$ is close to $p$ in the sliced Wasserstein-1 distance. Formally we have:
\begin{theorem}
\label{thm:dp}
Let $q$ be the output of \Cref{alg:dp}.
For any $\beta\in(0,1)$, with probability at least $1-\beta$ over the directions and Gaussian noise,
\begin{align}
\label{eq:dp_utility}
    \SW_1(p,q)
    \leq\frac{36}{k}
      +\frac4{\sqrt\beta}\left[\frac{\sigma^2(1+\log k)\log(6Lk^2)}{L}\right]^{1/4}
      +\frac{64}{\beta}\sqrt{\frac{d\log(6Lk^2)}{L}}.
\end{align}
\end{theorem}
The proof is given in \Cref{app:dp_utility}.
For $n\epsilon\geq1$ and $d<n\epsilon$, choose
\begin{align*}
    k=\lceil\sqrt{n\epsilon}\rceil,
    \qquad L=\lceil dn\epsilon(1+\log(n\epsilon))\rceil.
\end{align*}
Then $1+\log k=O(1+\log(n\epsilon))$ and, since $d<n\epsilon$, $\log(6Lk^2)=O(1+\log(n\epsilon))$.
Substituting these choices and $\sigma^2=8L(1+\log k)\log(1.25/\delta)/(n\epsilon)^2$ in \eqref{eq:dp_utility} gives
\begin{align}
\label{eq:dp_dimension_free_rate}
    \SW_1(p,q)\leq O_\beta\left(
       \frac{(1+\log(n\epsilon))^{3/4}\log^{1/4}(1.25/\delta)}{\sqrt{n\epsilon}}\right),
\end{align}
where the constant depends only on $\beta$. In particular, the bound does not depend on $d$.
In the remaining regimes, \eqref{eq:dp_dimension_free_rate} holds for any $k,L\geq1$: if $d\geq n\epsilon\geq1$, \Cref{lem:trivial_sw} gives $\SW_1(p,q)\leq2/\sqrt d\leq2/\sqrt{n\epsilon}$, and if $n\epsilon<1$, then $2 / \sqrt{n \epsilon} > 2\geq \SW_1(p, q)$, so \eqref{eq:dp_dimension_free_rate} holds with $\log(n \epsilon)$ read as $\max\{0, \log(n \epsilon)\}$.
This proves the $\tilde O(1/\sqrt{n\epsilon})$ guarantee stated in \Cref{sec:intro}.

\section{A Tractable Representation}
\label{sec:tractable}
Our main result from the previous section, \Cref{thm:dp}, says that it is information theoretically possible to return a private distribution $q$ such that $\SW_1(p,q) \leq \tilde{O}(1/\sqrt{n\epsilon})$. Indeed, $q$ can be obtained by postprocessing a set of sliced Chebyshev moments noised using the Gaussian mechanism. However, the theorem does not say anything about how to implement the postprocessing step efficiently, i.e. how to find $q$ that minimizes or approximately minimizes the moment error in \Cref{alg:dp}. We address that gap in this section, providing an effective heuristic for fitting a distribution $q$.

\paragraph{Mixture of product distributions.}
\label{sec:model}
Importantly, our approach requires a more efficient representation for $q$ than, for example, a naive $d$ dimensional histogram, which would require $2^{O(d)}$ parameters to even represent. We choose to instead optimize over a parameter-efficient mixture of product distributions. I.e., we consider distributions of the form $\qhat = \sum_{h=1}^{K} \pi_h \qhat_h$, where $K$ is an integer parameter, $\pi$ is a vector in the probability simplex, $\Delta_K$, and each $\qhat_h$ is a product distribution.

Concretely, for every index $a$ corresponding to a categorical variable with $r_a$ levels, the distribution of $\qhat_h$ at index $a$ is represented by a probability vector $\phi^{(h, a)} \in \Delta_{r_a}$. Each level is associated with a point on an equispaced grid of $r_a$ values on $[-1,1]$.
On the other hand, all continuous variables are modeled as a single joint Gaussian distribution with mean $\mu_h \in \R^{d_N}$ and covariance $\Sigma_h \in \R^{d_N\times d_N}$, where $d_N$ (respectively, $d_C$) counts the continuous (respectively, categorical) variables. We assume that $\Sigma_h$ is low-rank-plus-diagonal for some rank $R$, i.e. $\Sigma_h = U_h U_h^\top + \mathrm{diag}(D_h)$ with $U_h \in \R^{d_N \times R}$ and $D_h \in \R_{>0}^{d_N}$. The product of this single continuous distribution with the $d_C$ categorical distributions defines $\qhat_h$. Overall, we have:
\begin{align}
\label{eq:model}
   \qhat(x_C, x_N) = \sum_{h=1}^{K} \pi_h
   \paren{\prod_{a \in C} \phi^{(h, a)}(x_{a})}
   \cdot \mathcal{N}\paren{x_N; \mu_h, \Sigma_h}~.
\end{align}
Any distribution of this form requires $O(K(d_C r_{max} + d_N R))$ parameters to represent, where $r_{max}$ is an upper bound on the number of levels for any categorical variable.\footnote{It is worth noting that, in our implementation: (1) we rescale every column to $[-1,1]$ and measure moments of $\theta_\ell^\top x/\sqrt{d}$; as $[-1,1]^d/\sqrt{d}\subset\mathcal{B}^d$, the data lies in the unit ball as required by \Cref{alg:dp}, and the privacy analysis applies unchanged, and (2) the Gaussian block has unbounded support, so $\qhat$ is not itself supported on $\mathcal{B}^d$. We match moments using the unconstrained $\qhat$ with standard regularization techniques and clip samples coordinate-wise to $[-1,1]$ at the end, so the final private distribution is supported on the rescaled ball.}
We note that, for each component distribution, $\qhat_h$, categorical and continous variables are independent, but by mixing $K$ distributions, we can capture categorical-categorical and categorical-continuous dependence.

\paragraph{Optimization.}
With the model above in place, our goal becomes to fit a distribution that minimizes the moment error in \Cref{alg:dp}, i.e., to minimize the following expression over parameters of $\qhat$:
\begin{align}
    \label{what_we_minimize}
\frac1L\sum_{\ell=1}^L\sum_{j=1}^k\frac{(\tilde m_{j,\ell}-m_j(\theta_\ell,\qhat))^2}{j^2}.
\end{align}
While we cannot ensure that doing so obtains near-minimal error over all possible distributions, as will be shown in the next section, empirically we find that our mixture model is rich enough to accurately match many practical datasets from noisy sliced moments, even with $K = 5$ and $R = 2$.

We minimize \eqref{what_we_minimize} using gradient descent. We enforce simplex constraints on $\pi$ and each $\phi^{(h, a)}$ using a softmax reparameterization and we ensure the covariance diagonal is positive through a logarithmic reparameterization. To compute gradients, we take advantage of the fact that there is an efficient, differentiable procedure for computing $m_j(\theta_\ell,\qhat)$,
which we describe in \Cref{app:cheb_from_mixed_block}.

\begin{figure}[tbp]
  \centering
  \includegraphics[width=0.85\linewidth]{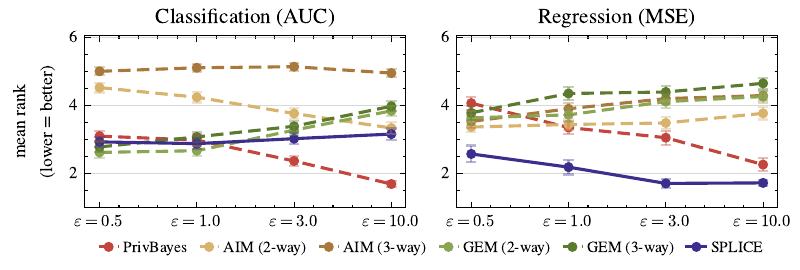}
  \caption{Mean TSTR rank across datasets (lower is better) by privacy budget, best downstream learner per dataset. Classification (AUC), $n = 94$ datasets, regression (MSE), $n = 67$.}
  \label{fig:mean_rank}
\end{figure}

\section{Experiments}
\label{sec:experiments}

\paragraph{Data and Baselines.}
We evaluate on 209 datasets from the TALENT benchmark \citep{JMLR:v26:25-0512}, spanning binary classification (75), multiclass classification (47), and regression (87), at privacy budgets $\eps \in \{0.5, 1, 3, 10\}$ with $\delta = 10^{-9}$ and five seeds per cell. Appendix~\ref{app:datasets} lists every dataset with its size, feature composition, and task. We consider a number of state-of-the-art baselines: AIM \citep{mckenna2022aim} (with 2-way and 3-way marginal workloads), PrivBayes \citep{PrivBayes:2017}, GEM \citep{liu2021iterative} (with 2-way and 3-way marginal workloads), and Private-GSD \citep{liu2023generating}, all under the public-schema convention described in \Cref{app:setup} for fairness. Our downstream utility for classification/regression measurements use the train-on-synthetic, test-on-real (TSTR) protocol, scoring each generated dataset by the best of four learners (logistic regression or ridge, random forest, XGBoost, TabPFN) and aggregating by rank. Despite checking 209 TALENT datasets, we \textit{only} report aggregate comparisons on what we call ``well-formed prediction tasks'' i.e. tasks where the train-on-real ceiling of AUC at least $0.7$ for classification and $R^2$ at least $0.3$ for regression. This means that ``unlearnable'' datasets do not contribute noisy wins; still, the vast majority of datasets are kept (184 of 209). Note that, unless otherwise specified, results for \methodname{} use the mixture of categorical-Gaussian product distributions of \Cref{alg:final}; a comparison among the \methodname{} model variants is in \Cref{app:variants}.

\begin{figure}[t]
    \centering
    \includegraphics[width=\linewidth]{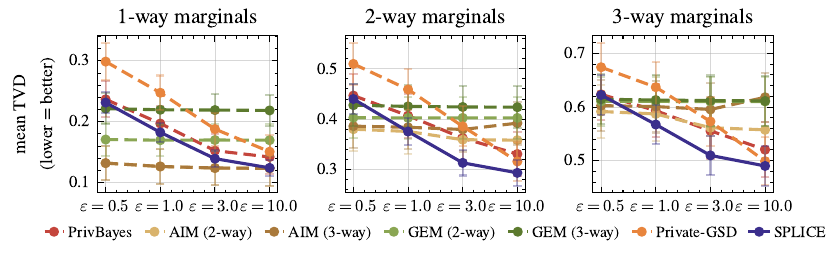}
    \caption{Mean $k$-way marginal total variation distance (lower is better) on a fixed set of $n = 43$ datasets where every method has a value at every budget. Left to right: 1-way, 2-way, 3-way marginals. \methodname{} never targets marginals yet leads 2-way and 3-way fidelity from $\eps \geq 1$.
}
    \label{fig:marginal_tvd}
\end{figure}

\paragraph{Overall comparison.}
\Cref{fig:overall} summarizes the benchmark at $\eps = 1$ by two composite ranks: distributional fidelity (1-way, 2-way, and 3-way marginal TVD, kernel MMD$^2$, and sliced Wasserstein-1) and downstream TSTR utility, macro-averaged over the classification and regression families with equal weight. \methodname{} is ranked first on \textit{four of the five} fidelity metrics, third on 1-way TVD behind only AIM (which measures those exact marginals directly), and outperforms every baseline on both composite axes simultaneously. On utility, \methodname{} dominates regression and is competitive on classification tasks. Per-metric values appear in \Cref{app:tables}.

\paragraph{Downstream predictive utility.}
\Cref{fig:mean_rank} shows mean TSTR rank by privacy budget. On regression, \methodname{} has the best mean rank at every budget (2.58, 2.19, 1.72, and 1.73 at $\eps = 0.5, 1, 3, 10$ over 67 datasets). On classification, \methodname{} is in the top 3 performers at each privacy budget. Classification is a more difficult setting for \methodname{}, as the target is discrete and the datasets have more categorical features on average; the strength of sliced Chebyshev moment projections is in natively handling continuous or ordinal values, while categorical values without order are not faithfully represented by the projection into continuous space. See \Cref{fig:rank_by_stratum} in the Appendix for the data type stratum breakdown.

\begin{figure}[tbp]
  \centering
  \begin{minipage}[t]{0.48\linewidth}
    \centering
    \includegraphics[width=\linewidth,height=1.8in,keepaspectratio]{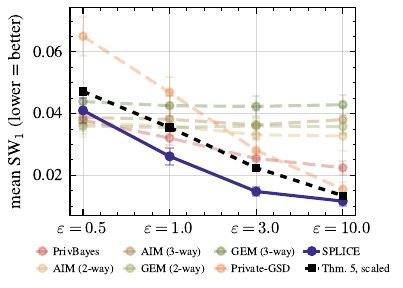}
    \caption{Sliced Wasserstein-1 distance (lower is better), 256 Monte Carlo directions on the unit ball, $n = 36$ datasets, with the bound of \Cref{thm:dp} parameterized at each dataset's $(n, d)$, minimized over $k$ and $L$, and divided through by a constant; the overlay compares rates in $\eps$, and baselines are faded.}
    \label{fig:sw_theory}
  \end{minipage}\hfill
  \begin{minipage}[t]{0.48\linewidth}
    \centering
    \includegraphics[width=\linewidth,height=1.8in,keepaspectratio]{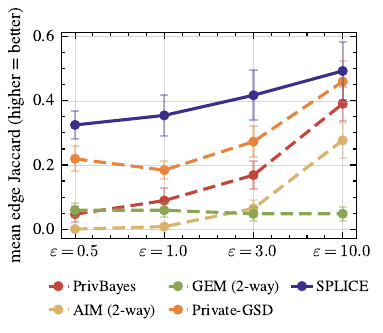}
    \caption{Jaccard similarity of undirected edge sets between the real-data and synthetic-data causal graphs (higher is better), $n = 11$ datasets. Error bars are $\pm 1$ SEM across datasets.}
    \label{fig:causal}
  \end{minipage}
\end{figure}

\paragraph{Marginal fidelity.} Despite never explicitly measuring a marginal, \methodname{} is \textbf{the best 3-way method at every budget, as well as the best 2-way method} from $\eps = 1$ (\Cref{fig:marginal_tvd}); AIM, which matches noisy 1-way marginals directly, is best on that metric, but even there its margin shrinks as $\eps$ increases. \methodname{}'s advantage grows with the interaction order, which is the main benefit of the method; matching higher-order Chebyshev moment measurements across random projections leads to distributional recovery containing higher-order interactions.

\paragraph{Distributional distance and causal structure.}
\methodname{} is theoretically motivated by the sliced Wasserstein-1 metric ($\SW_1$), which is \textit{distributional} and sensitive to higher order relationships in data. Perhaps unsurprisingly, our method is then best at every budget for $\SW_1$ (\Cref{fig:sw_theory}), and the method using a mixture of categorical-Gaussian product distributions matches the theoretical rate almost exactly (the worst-case rate is scaled by a large \textit{constant} factor, to match the practical $\SW_1$ error scale, but the decreasing rate across $\eps$ appears faithful). Additionally, the commonly used Kernel MMD distributional distance \citep{gretton2012kernel} gives the same ordering (\Cref{fig:mmd} in the appendix).

It has been noted that causal path analysis would be an impactful area for advances in DP synthetic data \citep{rosenblatt2024epistemic, ponomareva2025dp}. Thus, we run GRaSP, a standard causal discovery method \citep{lam2022greedy}, on real and synthetic data and compare the recovered graphs (\Cref{fig:causal}). At tight and moderate budgets \methodname{} roughly doubles edge recovery (measured F1) of the best marginal baseline (0.42 to 0.49 at $\eps \leq 3$ against 0.00 to 0.37).

\paragraph{Compute and reliability.}
Thanks to its near-independence from sample size and dimensionality by construction, \methodname{} runs in almost constant time and memory across three orders of magnitude of dataset size (30 to 120 seconds over a controlled grid from 12,500 to 400,000 rows and 6 to 24 columns, while e.g. Private-GSD times out beyond 25,000 rows, AIM struggles in higher dimensions and PrivBayes scales computationally with $n$ (\Cref{app:resources})). Additionally, across the full generations of the benchmark, \methodname{} \textbf{completed all 16,800 of its generations with zero failures}, while every other baseline failed somewhere due to each method's structural constraints.

\section{Conclusion}
\label{sec:conclusion}

We introduced \methodname{}, a differentially private synthetic data mechanism built on noisy sliced Chebyshev moments. Our analysis shows that matching moments along a number of random directions that is linear in the dimension controls the sliced Wasserstein-1 distance. Our mixture of categorical-Gaussian product distributions makes the mechanism practical, reliable and competitive: it outperforms strong marginal-based baselines on both composite axes simultaneously, distributional fidelity and downstream utility, across a 209-dataset benchmark, and in particular leads on regression, higher-order marginal fidelity and causal edge recovery.

\subsection*{Reproducibility}

\Cref{app:algorithm} states the full practical algorithm (\Cref{alg:final}) together with every default hyperparameter
(\Cref{tab:defaults}), and \Cref{app:setup} documents the experimental protocol, the public metadata convention,
and the baseline implementations with their sources.
\Cref{app:datasets} lists all benchmark datasets, and \Cref{app:tables} reports the per-metric values behind the figures.

An implementation of \methodname{} is available at \url{https://github.com/lurosenb/splice}. Please contact the authors for additional benchmark scripts and the complete results files.

\subsection*{Acknowledgements}
We thank Apoorv Vikram Singh for helpful discussions at the beginning of this project. This work was partially supported under the NSF AF program under award \#2427363. The work was also partially supported by a Google TPU Builders Award.

\bibliographystyle{references}
\bibliography{references}

\clearpage
\appendix

\section{Extended Related Work}
\label{app:related}

Here we expand the discussions of \Cref{sec:moment_matching} in the context of the surrounding literature.

\paragraph{Practical DP synthetic data.}
Most practical mechanisms for private tabular synthesis follow the Select-Measure-Estimate paradigm \citep{ponomareva2025dp}: choose a workload of low-dimensional statistics, measure them with calibrated noise, and fit a distribution to the noisy answers.
Model lineage is extensive, from early approaches like MWEM \citep{HardtR10, HardtLM12} to more recent graphical-model estimators \citep{mckenna2019graphical, mckenna2021winning, mckenna2022aim}, Bayesian networks \citep{PrivBayes:2017}, relaxed projection and generative estimators \citep{Aydore:2021, liu2021iterative, mckenna2021relaxed}, marginal-selection refinements \citep{zhang2021privsyn, cai2021data, liu2021leveraging, fuentes2024joint, mullins2024efficient}, and genetic search over discretized domains \citep{liu2023generating}.
These build on a long line of work on accurate query answering under DP, including consistency and matrix mechanism techniques \citep{Barak:2007, hay2010boosting, li2015matrix, mckenna2018optimizing, lee2015maximum, nikolov2013geometry}, and they have led to real deployments \citep{abowd2019census, TopDown, DP_heatmaps:2022}.
Because these mechanisms measure discrete marginals, they must discretize numerical features, a step the survey of \citet{ponomareva2025dp} identifies as a central difficulty for tabular data; \methodname{} sidesteps it by measuring moments of continuous projections directly.
A parallel line of work trains neural generative models under DP, from GANs \citep{jordon2018pate} to recent end-to-end approaches built on foundation models \citep{ponomareva2025dp}, although marginal-based methods still dominate tabular benchmarks \citep{rosenblatt2024data, Rosenblatt2020differentially}.

Closest to our release-once-then-fit structure within this line are private feature-sketch generators, which release a fixed noisy feature sketch and fit a generator to it: noisy sliced projections of records trained against smoothed-sliced $f$-divergences \citep{greenewald2024privacy}, random Fourier features for kernel MMD \citep{harder2021dpmerf}, Hermite polynomial features approximating a Gaussian kernel \citep{vinaroz2022hermite}, and compressive learning sketches \citep{chatalic2022compressive}.
\methodname{} shares the release-once-then-fit structure, but differs in two ways: its sliced Chebyshev sketch provably controls the distributional metric $\SW_1$, whereas these methods match kernel or feature discrepancies without a guarantee on a standard distributional metric; and these feature-matching generators have not demonstrated utility competitive with the marginal-based state of the art on realistic tabular workloads, which is the comparison our experiments target.

\paragraph{Hardness of private synthetic data.}
Answering all low-order statistics accurately is possible information-theoretically \citep{BlumLR13}, but generating synthetic data that does so is computationally hard in the worst case \citep{Dwork:2009, Ullman:2011, vadhan17complexity}, with sharp sample complexity transitions between pure and approximate DP \citep{steinke16between}.
This explains the field's split between practical heuristics and guarantee-first mechanisms, which can ignore these real-world constraints.

\paragraph{Distributional guarantees.}
A theoretical stream targets explicit statistical distance guarantees between the synthetic and true distributions.
For Wasserstein distance, private measures and random walks achieve the optimal rate, which unfortunately degrades as $n^{-1/d}$ in dimension $d$ \citep{Boedihardjo:2024, He:2023}, motivating work on low-dimensional structure \citep{He2023lowdim, Donhauser:2023}, online generation \citep{HeOnlineDP:2024}, instance optimality \citep{feldman2024instance}, and streaming moment estimation \citep{EpastoMMMVZ23}.
Weakening the metric helps: targeting smooth query classes improves the rates, although the sample complexity still depends exponentially on dimension \citep{WangEtAl:2016, DingHeZhu:2026}. Relative error guarantees also refine worst-case bounds \citep{GhaziPREM:2025, GKM21, DK22}.
Our approach descends from moment matching, where matching $k$ (Chebyshev) moments controls one-dimensional Wasserstein distance \citep{KongValiant:2017, BravermanKrishnanMusco:2022, musco2025sharper}; in this paper, we escape the dimension-exponential cost of the multivariate extension by matching moments only along random slices.

\paragraph{Sliced Wasserstein distance.}
Sliced Wasserstein distance was introduced as a tractable surrogate for Wasserstein distance \citep{rabin2011wasserstein, bonneel2015sliced} and is now a standard training objective and evaluation metric for generative models \citep{deshpande2018generative, deshpande2019max, kolouri2018sliced, kolouri2019sliced, kolouri2019generalized, liutkus2019sliced}.
Its statistical and geometric properties are an active topic, including sharp comparisons to the unsliced distance \citep{carlier2025sharp, nilesweed2026nearly}, kernel and intrinsic variants \citep{meunier2022distribution, rustamov2023intrinsic, bonet2023spherical}, and its role as an integral probability metric \citep{Muller:1997, Villani:2008}.
In private learning, sliced Wasserstein distance drives gradient flow generators \citep{sebag2025differentially} and trajectory synthesis \citep{gu2025private}.
To our knowledge, \methodname{} is the first to pair a sliced Wasserstein guarantee with a practical, state-of-the-art differentially private algorithm.

\section{Proof of the Moment Error Approximation Lemma}
\label{app:moment_matching}
Here we provide the deferred proof of \Cref{lem:error_matching}, which we restate below:
\begin{replemma}{lem:error_matching}
Fix any positive integer $k$ and any probability measure $p$ on
$\mathcal{B}^d$. Let $L$ be a positive integer, let $\beta\in(0,1)$, and let
$\theta_1,\ldots,\theta_L$ be independent draws from
$\Unif(S^{d-1})$. With probability at least $1-\beta$ over the choice of
$\theta_1,\ldots,\theta_L$, simultaneously for all probability measures $q$
on $\mathcal{B}^d$,
\begin{align*}
\tag{\ref{eq:direct_uniform_approximation}}
\left|
    \frac{1}{L}\sum_{\ell=1}^L \Gamma_k(\theta_\ell,p,q)
    - \E_{\theta\sim\Unif(S^{d-1})}\left[\Gamma_k(\theta,p,q)\right]
\right| \leq
\frac{32}{\beta}\sqrt{
    \frac{d\log(6Lk^2)}{L}
}.
\end{align*}
\end{replemma}

\begin{proof}[Proof of \Cref{lem:error_matching}]
Define
\begin{align*}
    Z \defeq
    \sup_q\left|
        \frac{1}{L}\sum_{\ell=1}^L \Gamma_k(\theta_\ell,p,q)
        - \E_{\theta\sim\Unif(S^{d-1})}\left[\Gamma_k(\theta,p,q)\right]
    \right|,
\end{align*}
where the supremum is over all probability measures on $\mathcal{B}^d$.
We bound $\E_\theta Z$ and apply Markov's inequality.
Expectation subscripts indicate the random family being averaged over;
all other random variables are held fixed.

\paragraph{Symmetrization and Gaussian Comparison.}
Let $\sigma_1,\ldots,\sigma_L$ be independent Rademacher random variables,
independent of the directions. A standard symmetrization inequality (see e.g. \citep[Lemma 2.3.1]{VanDerVaartWellner:1996}) gives
\begin{align}
\label{eq:direct_symmetrization}
    \E_\theta[Z]
    \leq
    \frac{2}{L}
    \E_{\theta,\sigma}
    \left[
        \sup_q\left|
            \sum_{\ell=1}^L \sigma_\ell \Gamma_k(\theta_\ell,p,q)
        \right|
    \right].
\end{align}
Fix the directions $\theta_1,\ldots,\theta_L$ and define
\begin{align}
\label{eq:direct_vector_definition}
    v_\ell(q)
    \defeq
    \left(
        \frac{m_1(\theta_\ell,p)-m_1(\theta_\ell,q)}{1},
        \ldots,
        \frac{m_k(\theta_\ell,p)-m_k(\theta_\ell,q)}{k}
    \right)\in\R^k,
\end{align}
so $\Gamma_k(\theta_\ell,p,q)=\|v_\ell(q)\|_2$.

Symmetry of the signs gives:
\begin{align}
    \label{eq:remove_abs}
\E_\sigma
\left[
    \sup_q\left|
        \sum_{\ell=1}^L \sigma_\ell\|v_\ell(q)\|_2
    \right|
\right]
&\leq
2\E_\sigma
\left[
    \sup_q
        \sum_{\ell=1}^L \sigma_\ell\|v_\ell(q)\|_2
\right].
\end{align}
For any realization of
$\sigma_1,\ldots,\sigma_L$,
\begin{align*}
    \sup_q\left|\sum_{\ell=1}^L\sigma_\ell\|v_\ell(q)\|_2\right|
    &\leq \sup_q\sum_{\ell=1}^L\sigma_\ell\|v_\ell(q)\|_2
        +\sup_q\sum_{\ell=1}^L(-\sigma_\ell)\|v_\ell(q)\|_2.
\end{align*}
Both suprema on the right are nonnegative, since we can always choose $q=p$,
in which case $v_\ell(q)=0$. Taking expectations and using the symmetry of
the random signs gives \eqref{eq:remove_abs}.

Gaussian comparison removes the $\ell_2$ norms:
\begin{align}
    \label{eq:final_gaussian_comparison}
    \E_\sigma\sup_q\sum_{\ell=1}^L\sigma_\ell\|v_\ell(q)\|_2
    \leq \sqrt{\pi/2} \E_g\sup_q\sum_{\ell=1}^L\langle g_\ell,v_\ell(q)\rangle,
\end{align}
Here $g_\ell=(g_{\ell1},\ldots,g_{\ell k})$ are independent standard Gaussian vectors. We apply the following lemma to each sign:

\begin{lemma}
\label{lem:gaussian_norm_comparison}
Fix $\ell\in\{1,\ldots,L\}$ and let $v_\ell(q)$ be as in \eqref{eq:direct_vector_definition}. Let $a(q)$ be a function over probability measures with
$\sup_q|a(q)|<\infty$. Let $\sigma$ be a Rademacher random variable and let
$g\sim\mathcal{N}(0,I_k)$. Then
\begin{align*}
    \E_\sigma\sup_q\{a(q)+\sigma\|v_\ell(q)\|_2\}
    \leq
    \E_g\sup_q\{a(q)+\sqrt{\pi/2}\langle g,v_\ell(q)\rangle\},
\end{align*}
where both suprema range over all probability measures $q$ on
$\mathcal{B}^d$.
\end{lemma}
\begin{proof}
Fix any two probability measures $q_+$ and $q_-$ on $\mathcal{B}^d$.
We restrict the Gaussian supremum to these two measures and use that, for two quantities $a$ and $b$, $\max\{a,b\} = \frac{a+b}{2} + \frac{|a-b|}{2}$ to obtain
\begin{align*}
    &\E_g\sup_q\{a(q)+\sqrt{\pi/2}\langle g,v_\ell(q)\rangle\}\\
    &\geq
    \E_g\max_{q\in\{q_+, q_-\}}\{a(q)+\sqrt{\pi/2}\langle g,v_\ell(q)\rangle\}\\
    &= \frac{a(q_+)+a(q_-)}{2}
        +\frac12\E_g\left|a(q_+)-a(q_-)
            +\sqrt{\pi/2}\langle g,v_\ell(q_+)-v_\ell(q_-)\rangle\right|\\
    &\geq \frac{a(q_+)+a(q_-)}{2}
        +\frac12\|v_\ell(q_+)-v_\ell(q_-)\|_2\\
    &\geq \frac{a(q_+)+a(q_-)}{2}
        +\frac{\|v_\ell(q_+)\|_2-\|v_\ell(q_-)\|_2}{2}.
\end{align*}
For the second inequality we use that, if $X$ is symmetric about zero, then
$\E|b+X|\geq\E|X|$.
Also, $\E|\langle g,u\rangle|=\sqrt{2/\pi}\|u\|_2$ for every $u$.
The last inequality follows from triangle inequality. Taking the supremum over
$q_+$ and $q_-$ separately on the right-hand side yields
\begin{align*}
    \frac12\sup_q\{a(q)+\|v_\ell(q)\|_2\}
    +\frac12\sup_q\{a(q)-\|v_\ell(q)\|_2\}
    =\E_\sigma\sup_q\{a(q)+\sigma\|v_\ell(q)\|_2\},
\end{align*}
which proves the claim.
\end{proof}

\eqref{eq:final_gaussian_comparison} is obtained by applying \Cref{lem:gaussian_norm_comparison} inductively to each $\sigma_\ell$,
conditioning on all other random variables and treating their contributions
as the offsets $a(q)$.

\paragraph{Concentration and Net Argument.} Combining \eqref{eq:remove_abs} and \eqref{eq:final_gaussian_comparison} gives:
\begin{align}
\label{eq:direct_vector_contraction}
&\E_\sigma
\left[
    \sup_q\left|
        \sum_{\ell=1}^L \sigma_\ell \Gamma_k(\theta_\ell,p,q)
    \right|
\right] \leq
\sqrt{2\pi}
\E_g
\left[
    \sup_q
    \sum_{\ell=1}^L\sum_{j=1}^k
        g_{\ell j}\frac{m_j(\theta_\ell,p)-m_j(\theta_\ell,q)}{j}
\right],
\end{align}
where the $g_{\ell j}$ are independent standard normal random variables.

For fixed directions and fixed $g_{\ell j}$, define
\begin{align}
\label{eq:direct_random_function}
    h(x)
    \defeq
    \sum_{\ell=1}^L\sum_{j=1}^k
        \frac{g_{\ell j}}{j}T_j(\theta_\ell^\top x),
    \qquad x\in\mathcal{B}^d.
\end{align}
By linearity of expectation, the expression inside the supremum on the
right-hand side of \eqref{eq:direct_vector_contraction} is
\begin{align*}
    \E_{x\sim p}[h(x)]-\E_{x\sim q}[h(x)].
\end{align*}
Since $q$ ranges over all probability measures on $\mathcal{B}^d$, we have:
\begin{align}
\label{eq:direct_distribution_to_point}
    \sup_q\left\{
        \E_{x\sim p}[h(x)]-\E_{x\sim q}[h(x)]
    \right\}
    = \E_{x\sim p}[h(x)]-\inf_{x\in\mathcal{B}^d}h(x) \leq
    \sup_{x\in\mathcal{B}^d}h(x)
    -\inf_{x\in\mathcal{B}^d}h(x).
\end{align}
It remains to bound the expected range of $h$ on $\mathcal B^d$. A net of this ball has log-cardinality linear in $d$.

Concretely, set $\rho=1/(Lk^2)$. By the standard volumetric covering bound for the
Euclidean ball (see, e.g., \citet{Vershynin:2026}), there is a $\rho$-net $\mathcal{N}_\rho\subseteq\mathcal{B}^d$
that satisfies
\begin{align}
\label{eq:direct_net_size}
    |\mathcal{N}_\rho|&\leq (6Lk^2)^d &&\text{and} &\forall x&\in\mathcal{B}^d,\quad \exists x'\in\mathcal{N}_\rho
    \text{ such that } \|x-x'\|_2\leq\rho.
\end{align}
For every fixed $x'\in\mathcal{N}_\rho$, the random variable
$h(x')$ is a centered Gaussian conditional on $\theta_1, \ldots, \theta_L$. Since $|T_j(t)|\leq 1$ for
$t\in[-1,1]$, its variance satisfies
\begin{align*}
    \operatorname{Var}_g(h(x')) = \sum_{\ell=1}^L\sum_{j=1}^k
    \frac{T_j(\theta_\ell^\top x')^2}{j^2}
    \leq
    L\sum_{j=1}^k\frac{1}{j^2}\leq \frac{\pi^2 L}{6}.
\end{align*}
A standard bound on the expected maximum of a collection of Gaussian variables \citep{Vershynin:2026} gives
\begin{align}
\label{eq:direct_net_maximum}
    \E_g
    \left[
        \max_{x'\in\mathcal{N}_\rho}h(x')
    \right]
    \leq
    \frac{\pi}{\sqrt{3}}\sqrt{L d\log(6Lk^2)}.
\end{align}

It remains to extend this bound to the entire ball. Let $x\in\mathcal{B}^d$ be some point in the ball and let $x'\in\mathcal{N}_\rho$ be the nearest point in our $\rho$-net.
For
$t\in[-1,1]$, we have $|T_j'(t)|\leq j^2$ \citep{Trefethen:2013}. It follows that:
\begin{align*}
    |h(x)-h(x')|
    =\left|\sum_{\ell=1}^L\sum_{j=1}^k
        \frac{g_{\ell j}}{j}
        \left(T_j(\theta_\ell^\top x)-T_j(\theta_\ell^\top x')\right)\right| &\leq\sum_{\ell=1}^L\sum_{j=1}^k
        \frac{|g_{\ell j}|}{j}
        \left|T_j(\theta_\ell^\top x)-T_j(\theta_\ell^\top x')\right|\\
    &\leq\sum_{\ell=1}^L\sum_{j=1}^k
        \frac{|g_{\ell j}|}{j} j^2
        |\theta_\ell^\top(x-x')|\\
&\leq \rho \sum_{\ell=1}^L\sum_{j=1}^k j|g_{\ell j}|
        .
\end{align*}
In the last step, we used Cauchy-Schwarz and the fact that $\|\theta_\ell\|_2 = 1$ and $\|x-x'\|_2 \leq \rho$.
It follows that:
\begin{align*}
    \E_g\sup_{x\in\mathcal{B}^d}h(x)
    \leq\E_g\max_{x'\in\mathcal{N}_\rho}h(x')
        +\rho\E_g\left[\sum_{\ell=1}^L\sum_{j=1}^k j|g_{\ell j}|\right]
    &\leq\E_g\max_{x'\in\mathcal{N}_\rho}h(x')+\rho \sqrt{\frac{2}{\pi}} L\frac{k(k+1)}{2}\\
    &\leq\E_g\max_{x'\in\mathcal{N}_\rho}h(x')+\rho Lk^2.
\end{align*}
Recalling that $\rho=1/(Lk^2)$, combining with
\eqref{eq:direct_net_maximum} yields
\begin{align}
\label{eq:direct_point_supremum}
    \E_g
    \left[
        \sup_{x\in\mathcal{B}^d}h(x)
    \right]
    \leq
    \frac{\pi}{\sqrt{3}}\sqrt{L d\log(6Lk^2)}+1.
\end{align}
The distribution of the $g_{\ell j}$ variables is unchanged when all signs are
negated, so $\E_g \left[\sup_{x\in\mathcal{B}^d}h(x)\right] = -\E_g \left[\inf_{x\in\mathcal{B}^d}h(x)\right]$. It follows that the right hand side of \eqref{eq:direct_distribution_to_point} can be bounded by twice the right-hand side of \eqref{eq:direct_point_supremum}. Combining
\eqref{eq:direct_symmetrization}, \eqref{eq:direct_vector_contraction},
\eqref{eq:direct_distribution_to_point}, and
\eqref{eq:direct_point_supremum}, we obtain
\begin{align}
\label{eq:direct_expectation_bound}
    \E_\theta[Z]
    \leq
    8\pi\sqrt{\frac{\pi}{6}}\sqrt{
        \frac{d\log(6Lk^2)}{L}
    }
    +\frac{8\sqrt{\pi/2}}{L}
    \leq 32\sqrt{\frac{d\log(6Lk^2)}{L}}.
\end{align}
The last inequality uses $d,k,L\geq 1$, so that
$1/L\leq\sqrt{d\log(6Lk^2)/L}$.

Markov's inequality completes the proof.
\end{proof}

\clearpage
\section{Proof of the Private Utility Theorem}
\label{app:dp_utility}
Here we bound the sliced Wasserstein-1 distance between the output of \Cref{alg:dp}, $q$, and a target distribution $p$. We could, by a direct analysis, bound the moment error of the differentially private, synthetic distribution $q$ by the norm of the added noise, which would grow with the number of directions $L$ (since $\sigma^2$ scales linearly in $L$). This, however, would force a trade-off against the direction-sampling error of \Cref{thm:average_control}, which decreases with $L$, and balancing the two leaves a $d^{1/4}$ factor in the rate. As an alternate strategy, we use the fact that $q$ minimizes the fitting objective to analyze how the noise correlates with the moment error of the distribution we fit for $q$. This correlation is controlled by a noise-weighted sum of re-centered Chebyshev polynomials $T_j(\theta_\ell^\top x)-T_j(0)$, where re-centering is free (the constant term cancels in the correlation) and ensures each summand is bounded by $O(j|\theta_\ell^\top x|)$. Examining this sum, we see it has variance proportional to $\lambda_{\max}\big(\frac{1}{L}\sum_\ell \theta_\ell\theta_\ell^\top\big) \approx 1/d$, which cleanly cancels out the factor of $d$ in the numerator from the net argument given in the proof of \Cref{lem:error_matching}.

We first record a bound that handles small $L$ and the regime $d\geq n\epsilon$.
\begin{lemma}
\label{lem:trivial_sw}
For any probability measures $p,q$ on $\mathcal{B}^d$, $\SW_1(p,q)\leq2/\sqrt d$.
\end{lemma}
\begin{proof}
Fix any coupling $(x,y)$ of $p$ and $q$. Its projected pushforward under $z \mapsto \theta^\top z$ is a coupling of $\bar p_\theta$ and $\bar q_\theta$, so $\SW_1(p,q) \leq \E_{x,y}\E_\theta|\theta^\top(x-y)|$.
By rotational invariance of $\Unif(\mathcal{S}^{d-1})$ and Jensen's inequality, for any $u\in\R^d$,
$\E_\theta| \theta^\top u| = \|u\|_2\E_\theta| \theta_1|\leq\|u\|_2\sqrt{\E_\theta\theta_1^2} = \|u\|_2/\sqrt d$. Since $\|x-y\|_2 \leq 2$ for $x, y \in \mathcal{B}^d$, we recover the claim.
\end{proof}
We restate \Cref{thm:dp} before proving it.
\begin{reptheorem}{thm:dp}
Let $q$ be the output of \Cref{alg:dp}.
For any $\beta\in(0,1)$, with probability at least $1-\beta$ over the directions and Gaussian noise,
\begin{align*}
\tag{\ref{eq:dp_utility}}
    \SW_1(p,q)
    \leq\frac{36}{k}
      +\frac4{\sqrt\beta}\left[\frac{\sigma^2(1+\log k)\log(6Lk^2)}{L}\right]^{1/4}
      +\frac{64}{\beta}\sqrt{\frac{d\log(6Lk^2)}{L}}.
\end{align*}
\end{reptheorem}
\begin{proof}
Fix the dataset and its empirical distribution $p$.
We first deal with the case $L<4d\log d$, which requires $d \geq 2$. In that case the last term of \eqref{eq:dp_utility} alone satisfies
\begin{align*}
    \frac{64}{\beta} \sqrt{\frac{d \log(6Lk^2)}{L}} \geq 64\sqrt{\frac{d \log 6}{4d \log d}} = 32\sqrt{\frac{\log 6}{\log d}} \geq \frac{2}{\sqrt d}~.
\end{align*}
By \Cref{lem:trivial_sw}, $\SW_1(p,q)\leq2/\sqrt d$, so \eqref{eq:dp_utility} holds with probability one. For the rest of the proof, assume $L\geq4d\log d$.
Let $\Phi$ denote the least-squares objective in \Cref{alg:dp}; a minimizer exists since probability measures on $\mathcal B^d$ are weakly compact and $\Phi$ is weakly continuous.
Since $p$ is feasible and $q$ minimizes this objective, $\Phi(q)\leq\Phi(p)$.
Use the noise $\eta_{j,\ell}\sim\mathcal N(0,j\sigma^2)$ from \Cref{alg:dp}. Here and below $\sum_{\ell,j}$ ranges over $1\leq\ell\leq L$, $1\leq j\leq k$.
Jensen's inequality gives
\begin{align*}
    \tilde\Gamma_{k,L}(p,q)
    \leq\left(\frac1L\sum_{\ell,j}\frac{(m_j(\theta_\ell, q) - m_j(\theta_\ell, p))^2}{j^2}\right)^{1/2}.
\end{align*}
Expanding $\Phi(q)\leq\Phi(p)$ gives:
\begin{align}
    \frac1L\sum_{\ell,j}\frac{(m_j(\theta_\ell, q) - m_j(\theta_\ell, p) - \eta_{j,\ell})^2}{j^2}
    &\leq\frac1L\sum_{\ell,j}\frac{\eta_{j,\ell}^2}{j^2}\notag\\
    \quad\Longrightarrow\quad
    \frac1L\sum_{\ell,j}\frac{(m_j(\theta_\ell, q) - m_j(\theta_\ell, p))^2}{j^2}
    &\leq\frac2L\sum_{\ell,j}
                       \frac{\eta_{j,\ell}(m_j(\theta_\ell, q) - m_j(\theta_\ell, p))}{j^2}.
\label{eq:centered_basic_inequality}
\end{align}
Define the noise weighted sum of re-centered Chebyshev polynomials, for $x \in \mathcal B^d$, as,
\begin{align}
\label{eq:dp_centered_polynomial}
    G(x)\defeq\frac{1}{L}\sum_{\ell,j}\frac{\eta_{j,\ell}}{j^2} [T_j(\theta_\ell^\top x)-T_j(0)]~.
\end{align}
The right side of \eqref{eq:centered_basic_inequality} equals
$2(\E_{q}G-\E_pG)$. Bounding $\E_{q}G$ by $\sup_xG(x)$ and using
$\E_\eta\E_pG=0$ therefore gives, conditional on the directions,
\begin{align}
\label{eq:centered_fit_to_supremum}
    \E_\eta \left[\frac{1}{L}\sum_{\ell,j} \frac{(m_j(\theta_\ell, q) - m_j(\theta_\ell, p))^2}{j^2}\right]\leq2\E_\eta\sup_{x\in\mathcal B^d}G(x).
\end{align}
This remains valid even though $q$ depends on the noise.

Let $S=L^{-1}\sum_{\ell=1}^L\theta_\ell\theta_\ell^\top$ be the
empirical second-moment matrix of the directions.
Since cosine is $1$-Lipschitz, the identity $T_j(u)=\cos(j\arccos u)$ gives
$|T_j(u)-T_j(0)|\leq j|\arccos u-\pi/2|=j|\arcsin u|$,
where we used $\arccos u-\pi/2=-\arcsin u$.
The elementary bound $|\arcsin u|\leq(\pi/2)|u|$ then gives
$|T_j(u)-T_j(0)|\leq(\pi/2)j|u|$ for $u\in[-1,1]$.
Fix $x\in\mathcal B^d$ and condition on the directions, so that the
coefficients of the noises in $G(x)$ are deterministic. Using $\operatorname{Var}(\eta_{j,\ell})=j\sigma^2$, we obtain,
\begin{align*}
    \operatorname{Var}_\eta G(x)
    &=\frac1{L^2}\sum_{\ell=1}^L\sum_{j=1}^k
       \frac{j\sigma^2}{j^4}
       [T_j(\theta_\ell^\top x)-T_j(0)]^2 =\frac{\sigma^2}{L^2}\sum_{\ell=1}^L\sum_{j=1}^k
       \frac{[T_j(\theta_\ell^\top x)-T_j(0)]^2}{j^3}\\
    &\leq\frac{\pi^2\sigma^2}{4L^2}\sum_{\ell=1}^L\sum_{j=1}^k
       \frac{(\theta_\ell^\top x)^2}{j} \leq \frac{\pi^2\sigma^2(1+\log k)}{4L}
       \left(\frac1L\sum_{\ell=1}^L(\theta_\ell^\top x)^2\right) \\
     &= \frac{\pi^2\sigma^2(1+\log k)}{4L}x^\top Sx~.
\end{align*}
The first inequality uses the Chebyshev bound above; the last line uses
$\sum_{j=1}^k1/j\leq1+\log k$ and the definition of $S$.
Since $S$ is positive semidefinite and $\|x\|_2\leq1$,
$x^\top Sx\leq\lambda_{\max}(S)\|x\|_2^2\leq\lambda_{\max}(S)$.
Thus the conditional variance is at most
$\pi^2\sigma^2(1+\log k)\lambda_{\max}(S)/(4L)$, uniformly over $x\in\mathcal B^d$.
For $d\geq2$, apply the matrix Bernstein expectation bound
\citep[Theorem~6.6.1, Eq.~(6.6.2)]{tropp2015introduction} to the independent,
centered matrices $X_\ell=\theta_\ell\theta_\ell^\top-I_d/d$.
Isotropy gives $\E_\theta X_\ell^2=(d-1)I_d/d^2$, and
$\lambda_{\max}(X_\ell)\leq1$, so
\begin{align*}
    \E_\theta\lambda_{\max}(S)
    \leq\frac1d+\sqrt{\frac{2\log d}{dL}}+\frac{\log d}{3L}
    \leq\frac2d.
\end{align*}
The last inequality uses $L\geq4d\log d$;
for $d=1$, $S=1$ deterministically.
This supplies the factor $1/d$ needed to cancel the
dimension dependence in the logarithm of the net size below.
Take a net $\mathcal N_\rho$ as in \eqref{eq:direct_net_size}, with
mesh $\rho=(Lk^2)^{-1}$ and size at most $(6Lk^2)^d$.
The Gaussian maximum bound, first conditional on the directions and then
averaged using Jensen's inequality, gives
\begin{align*}
    \E_{\theta,\eta}\left[\frac1L\sum_{\ell,j}\frac{(m_j(\theta_\ell, q) - m_j(\theta_\ell, p))^2}{j^2}\right]
    &\leq\frac{\pi\sigma}{\sqrt{2L}}\sqrt{(1+\log k)d\log(6Lk^2)}
                       \E_\theta\sqrt{\lambda_{\max}(S)}\\
    &\leq\pi\sigma\sqrt{\frac{(1+\log k)\log(6Lk^2)}{L}}.
\end{align*}
To extend this bound from the net to the ball, use $|T_j'|\leq j^2$
and the argument leading to \eqref{eq:direct_point_supremum}. The resulting error is at most
$\rho L^{-1}\sum_{\ell,j}\E|\eta_{j,\ell}|
\leq\frac{\sigma}{L}\sqrt{\frac2{\pi k}}$.
Thus \eqref{eq:centered_fit_to_supremum} yields
\begin{align}
    \E_{\theta,\eta}\left[\frac1L\sum_{\ell,j}\frac{(m_j(\theta_\ell, q) - m_j(\theta_\ell, p))^2}{j^2}\right]
    &\leq2\pi\sigma\sqrt{\frac{(1+\log k)\log(6Lk^2)}{L}}
       +\frac{2\sigma}{L}\sqrt{\frac2{\pi k}}\notag\\
    &\leq8\sigma\sqrt{\frac{(1+\log k)\log(6Lk^2)}{L}}.
\end{align}
Markov's inequality therefore gives, with probability at least $1-\beta/2$,
\begin{align*}
    \left(\frac1L\sum_{\ell,j}\frac{(m_j(\theta_\ell, q) - m_j(\theta_\ell, p))^2}{j^2}\right)^{1/2}\leq\frac4{\sqrt\beta}\left[\frac{\sigma^2(1+\log k)\log(6Lk^2)}{L}\right]^{1/4}.
\end{align*}
Combining this with \Cref{thm:average_control}, also at failure probability
$\beta/2$, proves that with probability at least $1-\beta$,
\begin{align*}
    \SW_1(p,q)\leq\frac{36}{k}
      +\frac4{\sqrt\beta}\left[\frac{\sigma^2(1+\log k)\log(6Lk^2)}{L}\right]^{1/4}
      +\frac{64}{\beta}\sqrt{\frac{d\log(6Lk^2)}{L}}.
\end{align*}
The simultaneous guarantee in \Cref{thm:average_control} applies to the fitted $q$, even though it depends on the directions and noise. This proves \eqref{eq:dp_utility}.
\end{proof}

\clearpage
\section{Chebyshev Moments for a Mixture of Categorical-Gaussian Product Distributions}
\label{app:cheb_from_mixed_block}
In this section we describe how to efficiently compute the sliced Chebyshev moment, $m_j(\theta, \qhat) = \E_{x\sim \qhat}[T_j(\theta^Tx)]$, for any direction $\theta \in \mathcal{S}^{d-1}$ and any distribution $\qhat$ that fits our
mixture of categorical-Gaussian product distributions from \Cref{sec:model}.

The computation is first simplified by passing to the standard moments. In particular, we can write the Chebyshev polynomial in the monomial basis as $T_j(y) = \sum_{t=0}^{j} c_{j, t} y^t$ and then use that:
\begin{align*}
\E_{x\sim \qhat}[T_j(\theta^Tx)] = \sum_{t=0}^{j} c_{j, t} \E_{x\sim \qhat}[(\theta^Tx)^t].
\end{align*}
We can further simplify the computation by splitting along the mixture components:
\begin{align}
\E_{x\sim\qhat}[T_j(\theta^T x)]
&= \sum_{h=1}^{K} \pi_h \sum_{t=0}^{j} c_{j,t}
    \E_{x\sim\qhat_h}[(\theta^T x)^t].
\end{align}

As in \Cref{sec:model}, let $C\subset \{1, \ldots, d\}$ contain the indices of all categorical variables in our data vectors. Let $N = \{1, \ldots, d\} \setminus C$ contain the indices of all continuous variables. Let $\theta_N$ and $x_N$ denote the vectors $\theta$ and $x$ restricted to the continuous indices. We have that:
\begin{align*}
\theta^Tx = \sum_{a \in C} \theta_a x_a + \theta_N^\top x_N.
\end{align*}
We can thus compute $\E_{x\sim\qhat_h}[(\theta^T x)^t]$ by iteratively applying the following standard expression, which holds for any independent random variables $A$ and $B$:
\begin{align}
   \E\big[(A + B)^t\big] = \sum_{s=0}^{t} \binom{t}{s} \E[A^s] \E[B^{t-s}],
\end{align}
In this case, the variables $\{x_a\}_{a \in C}$ and $x_N$ are jointly independent. We have thus reduced our task to evaluating:
\begin{align}
    &\E_{x\sim\qhat_h}\big[(\theta_a x_a)^t\big]
    \quad (a\in C) &
 & \text{and}&
    &\E_{x\sim\qhat_h}\big[(\theta_N^\top x_N)^t\big]
\end{align}
for all $t$ less than the maximum moment degree considered.

Since $x_a$ is a finite categorical variable, the first expectation reduces to a simple sum over the possible values of $x_a$. For $\E_{x\sim\qhat_h}\big[(\theta_N^\top x_N)^t\big]$, we use that, when $x\sim \qhat_h$, $\theta_N^\top x_N$ is a univariate Gaussian random variable with mean and variance:
\begin{align*}
\nu_h &= \theta_N^\top \mu_h & &\text{and} &\sigma_h^2 = \norm{U_h^\top \theta_N}_2^2 + \sum_{i \in N} D_{h, i} \theta_i^2.
\end{align*}
We can then just apply standard closed form expressions for the raw moments of a univariate Gaussian (see \cite{winkelbauer2012moments}).

We implemented the procedure described above in Torch, then used Autograd to compute gradients when optimizing over the parameters of $\hat{q}$.

\clearpage
\section{The \methodname{} Algorithm with a Mixture of Categorical-Gaussian Product Distributions}
\label{app:algorithm}

\Cref{alg:final} restates the mechanism of \Cref{sec:tractable} as we actually run it. The private release is exactly that of \Cref{alg:dp}; the ``Match'' step is carried out by fitting the mixture of categorical-Gaussian product distributions \Eqref{eq:model} using the exact, differentiable moment computation of \Cref{app:cheb_from_mixed_block}. \Cref{tab:defaults} lists the implementation defaults used in every experiment.

\begin{algorithm}[h]
\caption{\methodname{}: Sliced Moment Matching with a Mixture of Categorical-Gaussian Product Distributions}
\label{alg:final}
\begin{algorithmic}[1]
\Require Dataset $x_1, \dots, x_n$; privacy parameters $\eps, \delta > 0$; components $K$; covariance rank $R$; column split into blocks $C, N$.
\Ensure Synthetic distribution $\qhat$ of the form \Eqref{eq:model}.
\algrule
\State \textbf{Encode.} Normalize each continuous column to $[-1, 1]$ and map each categorical column's levels to equally spaced grid values in $[-1, 1]$, giving a design matrix $X \in [-1, 1]^{n \times d}$.
\State \textbf{Hyperparameters.} Set the maximum degree $k$ and number of directions $L$ (\Cref{tab:defaults}).
\State \textbf{Release.} Sample public directions $\theta_1, \dots, \theta_L \sim \Unif(S^{d-1})$, compute the scaled sliced Chebyshev moments of the projections $\theta_\ell^\top x_i / \sqrt{d}$, and add Gaussian noise calibrated to the $\ell_2$ sensitivity of the scaled moment vector via the analytic Gaussian mechanism.
\State \textbf{Fit.} Minimize the weighted squared moment error between the noisy targets and the exact model moments by gradient descent, with the softmax and logarithmic parameterizations, warm start, restarts, averaging, and loss from main body.
\State \textbf{Sample.} Draw components $h \sim \pi$, categorical levels from $\phi^{(h, a)}$, continuous blocks from $\mathcal{N}(\mu_h, \Sigma_h)$, and de-normalize each column to its original range.
\end{algorithmic}
\end{algorithm}

\begin{table}[h]
\centering
\begin{tabular}{lll}
\toprule
symbol & default & role \\
\midrule
$K$ & 5 & mixture components \\
$R$ & 2 & rank of each continuous covariance factor \\
$L$ & $\min(theory, 300)$ & number of directions \\
$k$ & $\min(theory, 30)$ & Chebyshev degree per direction \\
optimizer & Adam, cosine schedule & 2000 iterations, learning rate 0.02 \\
robustness & Huber loss ($\delta = 0.5$), EMA, restarts & stabilizes the non-convex fit \\
\bottomrule
\end{tabular}
\caption{Implementation defaults for \Cref{alg:final}, as used in all experiments.}
\label{tab:defaults}
\end{table}

\paragraph{More implementation details} \Cref{alg:final} inherits $(\eps, \delta)$-privacy by post-processing, with the scaled-moment sensitivity $\Delta_2 = \sqrt{\frac{4L}{n^2}\sum_{j=1}^k\frac1j}$ of \Cref{sec:splice} (we use the equivalent normalized basis $\Tbar_j = \sqrt{2/\pi} T_j$, with sensitivity $\sqrt{\frac{8L}{\pi n^2}\sum_{j=1}^k\frac1j}$). Continuous columns are \textit{not} rounded to a grid; a grid defines categorical values only. We calibrate $\sigma$ with the analytic Gaussian mechanism (as do all baselines, where appropriate, for fairness).

For our objective minimization we use gradient descent on the model parameters, enforcing the simplex constraints on $\pi$ and on each $\phi^{(h, a)}$ by a softmax reparameterization and keeping the covariance diagonal positive through a logarithmic parameterization.
We warm start the continuous-block means from the degree-one moments by least squares, since $m_1(\theta, \qhat)$ is linear in the means, and we stabilize the non-convex fit with random restarts (our experiments use one), an exponential moving average of the parameters \citep{polyak1992acceleration}, a Huber loss that caps the influence of the noisiest high-degree residuals \citep{huber1992robust}, and light entropy and ridge regularization. Note that all of these are standard moves, and additionally are post-processing with respect to the release of the noisy sliced Chebyshev moments.

\paragraph{Sampling.}
To draw a synthetic record we sample a component $h \sim \pi$, draw each categorical column independently from $\phi^{(h, a)}$, draw the continuous block from $\mathcal{N}(\mu_h, \Sigma_h)$, clip it to the encoded range, and de-normalize. We note that clipping is also what squares the model with \Cref{thm:dp}: the Gaussian block in \Eqref{eq:model} has unbounded support, whereas the theorem is stated for distributions on $\mathcal{B}^d$, so it is the clipped samples, rather than $\qhat$ itself, that satisfy the support assumption. Since fitting $\qhat$ and sampling from it are both post-processing of the noisy moments, privacy is unaffected either way. Finally, because the continuous block is Gaussian, samples are smooth and do not show the quantization artifacts of grid-based mechanisms.

\clearpage
\section{Additional Setup, Experiments and Ablations}
\label{app:setup}

\paragraph{Public metadata convention.} For a fair comparison, all methods are given the same kind of side information: column schema metadata is read off the training set and treated as public, so estimating it does not consume privacy budget. What each method reads differs. \methodname{} uses only per-column bounds (minimum and maximum) and the level set of each categorical column. Private-GSD takes an explicit dictionary of bounds; AIM treats every column as categorical and reads the categories from the data; GEM uses quantile bin edges; and PrivBayes builds its support from the observed values. Treating this metadata as public is a small idealization, but it is the standard one in the literature. We note that two of the baselines are given strictly more than bounds: GEM's quantile edges reveal the shape of each marginal, and AIM's category lists reveal exactly which values occur. \methodname{} has the smallest metadata footprint of the methods we compare.

\paragraph{Computational limits of baseline methods.}
Our comprehensive evaluation attempts $210 \times 4 \times 5 = 4200$ generation tasks per algorithm (incorporating the 209 benchmark datasets along with one synthetic smoke-test dataset excluded from the final metrics). We enforce a strict two-hour timeout per task (all methods were GPU accelerated, except for PrivBayes which has a fast C implementation). PrivBayes fails on 34 tasks due to the most massive datasets (its runtime scales as $\approx n^2$, and our estimated runtime for these larger datasets was multiple days for the method). AIM with 2-way workloads fails on 34 tasks due to timeouts and GPU memory exhaustion, primarily at $\eps = 10$, where it allocates larger budgets across more measurement iterations; its 3-way variant drops 153 tasks across ten high-dimensional datasets, seven of which fail across all budgets regardless of row count. GEM utilizing 3-way workloads drops 50 tasks on wide datasets with high cardinality, whereas its 2-way version successfully completes all tasks. Private-GSD runs out of host memory on larger tables and is thus evaluated solely on the subsets in general (this is despite being run on an NVIDIA RTX PRO 6000 Blackwell GPU with 96 GB of VRAM). Conversely, \methodname{} successfully executed all 16,800 of its generations, with no timeouts or out-of-memory errors. This reliability is a result of the method architecture: the sliced release compresses the data into fixed-size moment sketches, scaling linearly with $n$ and avoiding the combinatorial explosion associated with width, cardinality, or privacy budget.

\paragraph{Baseline implementations and parameter settings.} We use the maintained or author-released implementation of every baseline. AIM runs through the OpenDP Library (\url{https://github.com/opendp/opendp}, version 0.14.2), whose implementation builds on the graphical-model inference of private-pgm (\url{https://github.com/ryan112358/private-pgm}, \texttt{mbi} 1.1.0). 

GEM uses the authors' \texttt{dp-query-release} codebase (\url{https://github.com/terranceliu/dp-query-release}), consolidated into our harness without algorithmic changes. PrivBayes runs the authors' reference C++ implementation through a wrapper following SDGym (\url{https://github.com/sdv-dev/SDGym}). Private-GSD uses the authors' implementation (\url{https://github.com/giusevtr/private_gsd}). Causal discovery uses the GRaSP implementation in \texttt{causal-learn} (\url{https://github.com/py-why/causal-learn}). Every method runs with the settings recommended by its source implementation, and no method (including \methodname{}, whose $k$ and $L$ follow the fixed formulas of \Cref{tab:defaults}) receives per-dataset hyperparameter tuning. The only per-baseline variation we report is the workload degree (2-way and 3-way marginals for AIM and GEM), shown as separate baselines, and all methods share the privacy accounting conventions and public metadata described above.

\begin{figure}[h]
  \centering
  \includegraphics[width=\linewidth]{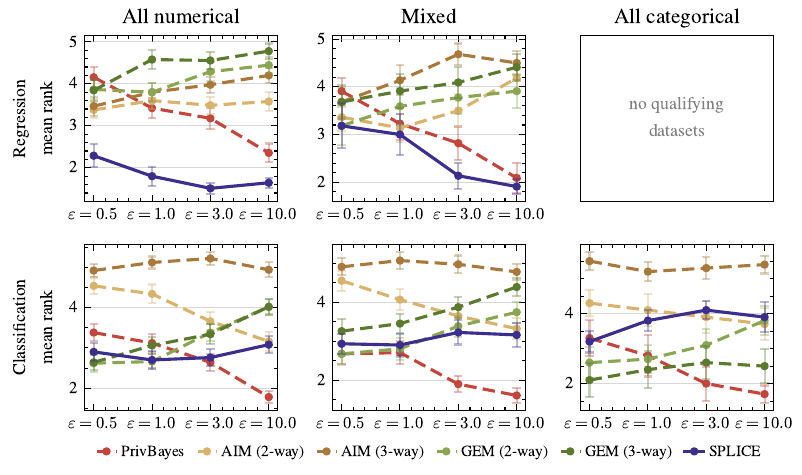}
  \caption{
Mean TSTR rank (lower is better) by privacy budget, split by task family (top: regression, bottom: classification) and by feature type (columns: all numerical, mixed, all categorical). Each panel uses a fixed set of well-formed datasets on which every plotted method has a value at every budget: $n = 45$ (numerical) and $22$ (mixed) for regression; $n = 50$, $31$, and $10$ for classification. Error bars are $\pm 1$ SEM across datasets. The benchmark has no well-formed all-categorical regression datasets.
  }
  \label{fig:rank_by_stratum}
\end{figure}

\begin{figure}[h]
  \centering
  \includegraphics[width=0.48\linewidth]{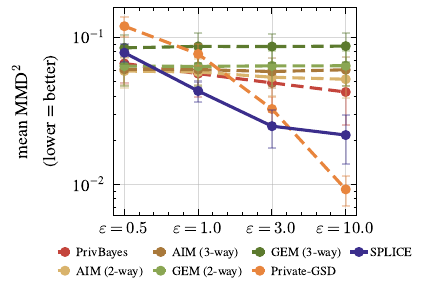}
  \caption{
Unbiased RBF-kernel MMD$^2$ between real and synthetic data (lower is better, log scale), median-heuristic bandwidth, $n = 36$ datasets. Error bars are $\pm 1$ SEM across datasets.
  }
  \label{fig:mmd}
\end{figure}

\subsection{\methodname{} Model Variants}
\label{app:variants}

\begin{figure}[h]
  \centering
  \includegraphics[width=\linewidth]{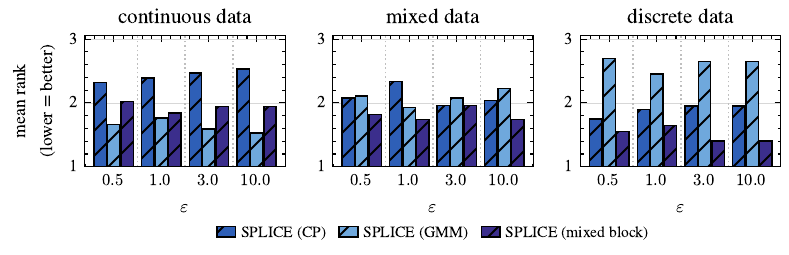}
  \caption{
Mean rank (lower is better) of the three \methodname{} variants (CP, GMM, and the mixture of categorical-Gaussian product distributions, labeled ``mixed block''), split by the fraction of discrete columns: continuous ($\leq 0.3$), mixed, and discrete ($\geq 0.7$). Left to right, the strata contain $n = 50$ to $51$, $27$, and $20$ datasets per budget.
  }
  \label{fig:variants}
\end{figure}

The mixture of categorical-Gaussian product distributions from \Cref{sec:tractable} generalizes two simpler models: a CP-style mixture of categorical product distributions, and a low-rank Gaussian mixture. (Figure legends in this appendix abbreviate the full model as ``mixed block''.) We also ran both of these special cases as \methodname{} variants, fit to the exact same noisy moments. \Cref{fig:variants} ranks the three on roughly 97 datasets, stratified by the fraction of discrete columns. The results are what the model classes suggest: the Gaussian mixture is best on continuous data, the CP model is best on discrete data (where the Gaussian mixture does poorly), and the full mixed model is best on mixed data.

\begin{figure}[h]
  \centering
  \includegraphics[width=\linewidth]{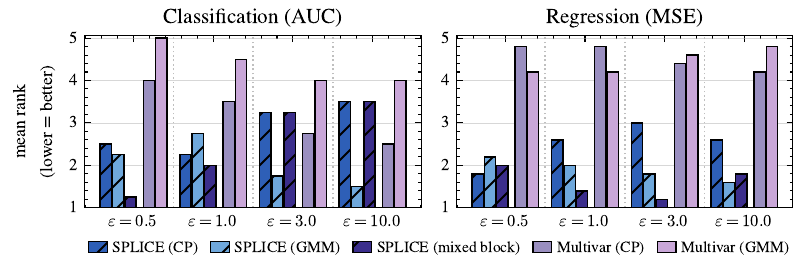}
  \caption{
Mean rank (lower is better) on 4-dimensional datasets (each restricted to its top 3 features), where the multivariate Chebyshev mechanism is still computationally feasible: the two multivariate variants against the sliced \methodname{} variants. Left: classification, $n = 4$ datasets. Right: regression, $n = 5$.
  }
  \label{fig:arena}
\end{figure}

\Cref{fig:arena} compares slicing to the multivariate Chebyshev mechanism of \cite{musco2025sharper} on 4-dimensional datasets, the only regime where the latter is computationally feasible. A sliced \methodname{} variant is ranked first for every task type and budget. The multivariate variants are behind on regression at every $\eps$; multivariate CP reaches second on classification at the looser budgets but is never first. Against the full set of baselines, the multivariate variants rank at or near the bottom. In other words, slicing helps even when $d$ is small enough that the curse of dimensionality is not binding.

\clearpage
\section{Result Tables}
\label{app:tables}

This section gives the numbers behind the figures in the main text. Table~\ref{tab:fidelity_eps1} reports the five fidelity metrics at $\eps = 1$, i.e., the components of the fidelity axis of \Cref{fig:overall}. Table~\ref{tab:fidelity_rank} gives the mean fidelity rank at every budget. Table~\ref{tab:tstr_rank} gives mean TSTR rank by budget and task family on a fixed set of datasets, and Table~\ref{tab:tstr_learner} breaks the $\eps = 1$ column down by downstream learner. Every table is computed from the same saved results and the same dataset filters as the figure it supports.

\begin{table}[h]
\centering
\small
\setlength{\tabcolsep}{4pt}
\resizebox{\linewidth}{!}{
\begin{tabular}{lccccc}
\toprule
Method & 1-way TVD ($n=58$) & 2-way TVD ($n=58$) & 3-way TVD ($n=58$) & MMD$^2$ ($n=40$) & SW$_1$ ($n=40$) \\
\midrule
PrivBayes & 0.214 {\scriptsize$\pm$0.027} & 0.442 {\scriptsize$\pm$0.037} & 0.642 {\scriptsize$\pm$0.041} & 0.057 {\scriptsize$\pm$0.016} & 0.032 {\scriptsize$\pm$0.002} \\
AIM (2-way) & \textbf{0.143} {\scriptsize$\pm$0.028} & 0.412 {\scriptsize$\pm$0.038} & 0.638 {\scriptsize$\pm$0.042} & 0.058 {\scriptsize$\pm$0.012} & 0.036 {\scriptsize$\pm$0.004} \\
AIM (3-way) & \textbf{0.143} {\scriptsize$\pm$0.028} & 0.420 {\scriptsize$\pm$0.038} & 0.650 {\scriptsize$\pm$0.040} & 0.061 {\scriptsize$\pm$0.012} & 0.039 {\scriptsize$\pm$0.004} \\
GEM (2-way) & 0.189 {\scriptsize$\pm$0.026} & 0.438 {\scriptsize$\pm$0.036} & 0.658 {\scriptsize$\pm$0.039} & 0.065 {\scriptsize$\pm$0.016} & 0.036 {\scriptsize$\pm$0.002} \\
GEM (3-way) & 0.243 {\scriptsize$\pm$0.025} & 0.462 {\scriptsize$\pm$0.036} & 0.662 {\scriptsize$\pm$0.039} & 0.088 {\scriptsize$\pm$0.018} & 0.045 {\scriptsize$\pm$0.003} \\
Private-GSD & 0.270 {\scriptsize$\pm$0.027} & 0.486 {\scriptsize$\pm$0.035} & 0.668 {\scriptsize$\pm$0.038} & 0.077 {\scriptsize$\pm$0.013} & 0.047 {\scriptsize$\pm$0.004} \\
Sliced (mixed block, ours) & 0.187 {\scriptsize$\pm$0.013} & \textbf{0.388} {\scriptsize$\pm$0.023} & \textbf{0.586} {\scriptsize$\pm$0.030} & \textbf{0.043} {\scriptsize$\pm$0.006} & \textbf{0.026} {\scriptsize$\pm$0.003} \\
\bottomrule
\end{tabular}
}
\caption{Distributional fidelity at $\varepsilon = 1$, the five components of the Figure~\ref{fig:overall} fidelity axis.
Each entry is the mean $\pm 1$ SEM over the datasets ($n$ per column) where every method has a value; lower is better and the best value per column is bold.}
\label{tab:fidelity_eps1}
\end{table}

\begin{table}[h]
\centering
\small
\setlength{\tabcolsep}{4pt}
\begin{tabular}{lcccc}
\toprule
Method & $\varepsilon = 0.5$ & $\varepsilon = 1.0$ & $\varepsilon = 3.0$ & $\varepsilon = 10.0$ \\
\midrule
PrivBayes & 5.00 & 3.40 & 3.00 & 3.20 \\
AIM (2-way) & \textbf{1.40} & 2.40 & 3.00 & 3.40 \\
AIM (3-way) & 2.60 & 3.60 & 4.60 & 5.00 \\
GEM (2-way) & 2.80 & 4.20 & 5.60 & 5.60 \\
GEM (3-way) & 5.40 & 6.20 & 7.00 & 6.80 \\
Private-GSD & 7.00 & 6.80 & 3.40 & 2.40 \\
Sliced (mixed block, ours) & 3.80 & \textbf{1.40} & \textbf{1.40} & \textbf{1.60} \\
\bottomrule
\end{tabular}
\caption{Mean fidelity rank (over the five distributional metrics of Table~\ref{tab:fidelity_eps1}) at every privacy budget.
Rank 1 is best out of 7; the best value per column is bold.}
\label{tab:fidelity_rank}
\end{table}

\begin{table}[h]
\centering
\small
\setlength{\tabcolsep}{4pt}
\resizebox{\linewidth}{!}{
\begin{tabular}{lcccccccc}
\toprule
& \multicolumn{4}{c}{Classification ($n=91$)} & \multicolumn{4}{c}{Regression ($n=67$)} \\
\cmidrule(lr){2-5}\cmidrule(lr){6-9}
Method & $\varepsilon=0.5$ & $\varepsilon=1.0$ & $\varepsilon=3.0$ & $\varepsilon=10.0$ & $\varepsilon=0.5$ & $\varepsilon=1.0$ & $\varepsilon=3.0$ & $\varepsilon=10.0$ \\
\midrule
PrivBayes & 3.13 {\scriptsize$\pm$0.16} & 2.95 {\scriptsize$\pm$0.17} & \textbf{2.32} {\scriptsize$\pm$0.15} & \textbf{1.71} {\scriptsize$\pm$0.11} & 4.07 {\scriptsize$\pm$0.19} & 3.36 {\scriptsize$\pm$0.19} & 3.06 {\scriptsize$\pm$0.21} & 2.27 {\scriptsize$\pm$0.18} \\
AIM (2-way) & 4.52 {\scriptsize$\pm$0.15} & 4.22 {\scriptsize$\pm$0.17} & 3.68 {\scriptsize$\pm$0.16} & 3.27 {\scriptsize$\pm$0.17} & 3.37 {\scriptsize$\pm$0.16} & 3.45 {\scriptsize$\pm$0.17} & 3.49 {\scriptsize$\pm$0.18} & 3.78 {\scriptsize$\pm$0.18} \\
AIM (3-way) & 4.98 {\scriptsize$\pm$0.12} & 5.11 {\scriptsize$\pm$0.11} & 5.14 {\scriptsize$\pm$0.12} & 4.93 {\scriptsize$\pm$0.13} & 3.54 {\scriptsize$\pm$0.19} & 3.91 {\scriptsize$\pm$0.18} & 4.21 {\scriptsize$\pm$0.15} & 4.30 {\scriptsize$\pm$0.15} \\
GEM (2-way) & \textbf{2.64} {\scriptsize$\pm$0.16} & \textbf{2.71} {\scriptsize$\pm$0.15} & 3.35 {\scriptsize$\pm$0.16} & 3.90 {\scriptsize$\pm$0.15} & 3.64 {\scriptsize$\pm$0.21} & 3.73 {\scriptsize$\pm$0.20} & 4.12 {\scriptsize$\pm$0.19} & 4.27 {\scriptsize$\pm$0.18} \\
GEM (3-way) & 2.79 {\scriptsize$\pm$0.17} & 3.12 {\scriptsize$\pm$0.16} & 3.44 {\scriptsize$\pm$0.17} & 3.98 {\scriptsize$\pm$0.15} & 3.79 {\scriptsize$\pm$0.20} & 4.36 {\scriptsize$\pm$0.19} & 4.40 {\scriptsize$\pm$0.18} & 4.66 {\scriptsize$\pm$0.16} \\
Sliced (mixed block, ours) & 2.95 {\scriptsize$\pm$0.16} & 2.89 {\scriptsize$\pm$0.16} & 3.07 {\scriptsize$\pm$0.17} & 3.20 {\scriptsize$\pm$0.17} & \textbf{2.58} {\scriptsize$\pm$0.25} & \textbf{2.19} {\scriptsize$\pm$0.22} & \textbf{1.72} {\scriptsize$\pm$0.13} & \textbf{1.73} {\scriptsize$\pm$0.10} \\
\bottomrule
\end{tabular}
}
\caption{Downstream TSTR mean rank ($\pm 1$ SEM across datasets) by privacy budget and task family, best downstream learner per dataset.
Datasets are the fixed sets of well-formed tasks covered by every shown method at every budget, so the composition is identical across columns; rank 1 is best out of 6 and the best value per column is bold.
Private-GSD is excluded because it covers only 10 of the 73 well-formed regression datasets.}
\label{tab:tstr_rank}
\end{table}

\begin{table}[h]
\centering
\small
\setlength{\tabcolsep}{4pt}
\resizebox{\linewidth}{!}{
\begin{tabular}{lcccccccc}
\toprule
& \multicolumn{4}{c}{Classification} & \multicolumn{4}{c}{Regression} \\
\cmidrule(lr){2-5}\cmidrule(lr){6-9}
Method & LogReg/Ridge & RF & XGB & TabPFN & LogReg/Ridge & RF & XGB & TabPFN \\
\midrule
PrivBayes & 3.02 & \textbf{2.56} & \textbf{2.71} & 2.85 & 3.27 & 3.04 & 3.18 & 3.40 \\
AIM (2-way) & 4.24 & 4.08 & 4.03 & 4.14 & 3.43 & 3.96 & 3.97 & 3.34 \\
AIM (3-way) & 4.92 & 4.81 & 4.80 & 4.92 & 3.70 & 3.96 & 4.03 & 3.76 \\
GEM (2-way) & 3.15 & 3.31 & 3.19 & 3.16 & 4.16 & 3.91 & 3.94 & 3.75 \\
GEM (3-way) & 3.13 & 3.41 & 3.36 & 3.10 & 4.33 & 4.10 & 4.06 & 4.43 \\
Sliced (mixed block, ours) & \textbf{2.53} & 2.84 & 2.90 & \textbf{2.83} & \textbf{2.10} & \textbf{2.03} & \textbf{1.82} & \textbf{2.31} \\
\bottomrule
\end{tabular}
}
\caption{Downstream TSTR mean rank at $\varepsilon = 1$ broken out by downstream learner (the linear column is logistic regression on classification and ridge on regression).
Same fixed dataset sets and rank convention as Table~\ref{tab:tstr_rank}; the headline figures use the best learner per dataset.}
\label{tab:tstr_learner}
\end{table}

\clearpage
\section{Dataset details}
\label{app:datasets}
\begingroup
\footnotesize
\setlength{\tabcolsep}{4pt}
\begin{longtable}{llrrrrc}
\caption{The 209 TALENT datasets in the benchmark: task type, training and test rows, numerical and categorical feature counts, and whether the dataset enters the well-formed aggregate set (WF; real-data ceiling AUC $\geq 0.7$ or $R^2 \geq 0.3$). A few long dataset names are truncated.}\label{tab:datasets} \\
\toprule
Dataset & Task & $n_{\mathrm{train}}$ & $n_{\mathrm{test}}$ & num. & cat. & WF \\
\midrule
\endfirsthead
\multicolumn{7}{l}{\footnotesize Table \ref{tab:datasets} continued.} \\
\toprule
Dataset & Task & $n_{\mathrm{train}}$ & $n_{\mathrm{test}}$ & num. & cat. & WF \\
\midrule
\endhead
\bottomrule
\endlastfoot
1000-Cameras-Dataset & regression & 830 & 208 & 10 & 0 & \checkmark \\
2dplanes & regression & 32,614 & 8,154 & 10 & 0 & \checkmark \\
3D\_\allowbreak{}Estimation\_\allowbreak{}using\_\allowbreak{}RSSI\_\allowbreak{}of\_\allowbreak{}WLAN\_\allowbreak{}dataset & regression & 4,608 & 1,152 & 6 & 0 & \checkmark \\
3D\_\allowbreak{}Estimation\_\allowbreak{}using\_\allowbreak{}RSSI\_\allowbreak{}of\_\allowbreak{}WLAN\_\allowbreak{}dataset\_\allowbreak{}c... & regression & 11,520 & 2,880 & 12 & 0 & \checkmark \\
abalone & multiclass & 3,341 & 836 & 7 & 1 & \checkmark \\
Abalone\_\allowbreak{}reg & regression & 3,341 & 836 & 7 & 1 & \checkmark \\
accelerometer & multiclass & 122,403 & 30,601 & 3 & 1 & \checkmark \\
ada\_\allowbreak{}prior & binary & 3,649 & 913 & 6 & 8 & \checkmark \\
airfoil\_\allowbreak{}self\_\allowbreak{}noise & regression & 1,202 & 301 & 5 & 0 & \checkmark \\
airlines\_\allowbreak{}seed\_\allowbreak{}0\_\allowbreak{}nrows\_\allowbreak{}2000\_\allowbreak{}nclasses\_\allowbreak{}10\_\allowbreak{}nco... & binary & 1,600 & 400 & 3 & 4 &  \\
Amazon\_\allowbreak{}employee\_\allowbreak{}access & binary & 26,215 & 6,554 & 0 & 7 & \checkmark \\
analcatdata\_\allowbreak{}supreme & regression & 3,241 & 811 & 7 & 0 & \checkmark \\
Another-Dataset-on-used-Fiat-500-(1538-rows) & regression & 1,230 & 308 & 6 & 0 & \checkmark \\
archive2 & regression & 914 & 229 & 11 & 1 & \checkmark \\
artificial-characters & multiclass & 8,174 & 2,044 & 7 & 0 & \checkmark \\
auction\_\allowbreak{}verification & regression & 1,634 & 409 & 6 & 1 & \checkmark \\
autoUniv-au7-1100 & multiclass & 880 & 220 & 8 & 4 &  \\
avocado\_\allowbreak{}sales & regression & 14,599 & 3,650 & 11 & 2 & \checkmark \\
bank & binary & 36,168 & 9,043 & 7 & 9 & \checkmark \\
bank8FM & regression & 6,553 & 1,639 & 8 & 0 & \checkmark \\
Bank\_\allowbreak{}Customer\_\allowbreak{}Churn\_\allowbreak{}Dataset & binary & 8,000 & 2,000 & 6 & 4 & \checkmark \\
banknote\_\allowbreak{}authentication & binary & 1,097 & 275 & 4 & 0 &  \\
baseball & multiclass & 1,072 & 268 & 15 & 1 & \checkmark \\
Basketball\_\allowbreak{}c & binary & 1,072 & 268 & 11 & 0 & \checkmark \\
Bias\_\allowbreak{}correction\_\allowbreak{}r & regression & 6,180 & 1,545 & 21 & 0 & \checkmark \\
Bias\_\allowbreak{}correction\_\allowbreak{}r\_\allowbreak{}2 & regression & 6,180 & 1,545 & 21 & 0 & \checkmark \\
BLE\_\allowbreak{}RSSI\_\allowbreak{}dataset\_\allowbreak{}for\_\allowbreak{}Indoor\_\allowbreak{}localization & multiclass & 7,987 & 1,997 & 3 & 0 & \checkmark \\
BNG(breast-w) & binary & 31,492 & 7,874 & 9 & 0 & \checkmark \\
BNG(cmc) & multiclass & 44,236 & 11,060 & 2 & 7 & \checkmark \\
BNG(echoMonths) & regression & 13,996 & 3,500 & 5 & 3 & \checkmark \\
BNG(lowbwt) & regression & 24,883 & 6,221 & 2 & 7 & \checkmark \\
BNG(mv) & regression & 62,985 & 15,747 & 7 & 3 & \checkmark \\
BNG(stock) & regression & 47,239 & 11,810 & 9 & 0 & \checkmark \\
BNG(tic-tac-toe) & binary & 31,492 & 7,874 & 0 & 9 & \checkmark \\
Brazilian\_\allowbreak{}houses\_\allowbreak{}reproduced & regression & 8,553 & 2,139 & 8 & 0 & \checkmark \\
California-Housing-Classification & binary & 16,512 & 4,128 & 8 & 0 & \checkmark \\
car-evaluation & multiclass & 1,382 & 346 & 0 & 21 & \checkmark \\
Cardiovascular-Disease-dataset & binary & 56,000 & 14,000 & 5 & 6 & \checkmark \\
CDC\_\allowbreak{}Diabetes\_\allowbreak{}Health\_\allowbreak{}Indicators & binary & 202,944 & 50,736 & 7 & 14 & \checkmark \\
churn & binary & 4,000 & 1,000 & 16 & 4 & \checkmark \\
Click\_\allowbreak{}prediction\_\allowbreak{}small & binary & 31,958 & 7,990 & 3 & 0 &  \\
cmc & multiclass & 1,178 & 295 & 2 & 7 & \checkmark \\
combined\_\allowbreak{}cycle\_\allowbreak{}power\_\allowbreak{}plant & regression & 7,654 & 1,914 & 4 & 0 & \checkmark \\
compass & binary & 13,315 & 3,329 & 8 & 9 & \checkmark \\
compass\_\allowbreak{}reg & regression & 13,315 & 3,329 & 17 & 0 & \checkmark \\
concrete\_\allowbreak{}compressive\_\allowbreak{}strength & regression & 824 & 206 & 8 & 0 & \checkmark \\
contraceptive\_\allowbreak{}method\_\allowbreak{}choice & multiclass & 1,178 & 295 & 5 & 4 & \checkmark \\
CookbookReviews & regression & 14,545 & 3,637 & 7 & 0 &  \\
CPMP-2015-runtime-regression & regression & 1,686 & 422 & 22 & 1 & \checkmark \\
CPS1988 & regression & 22,524 & 5,631 & 2 & 4 &  \\
cpu\_\allowbreak{}act & regression & 6,553 & 1,639 & 21 & 0 & \checkmark \\
cpu\_\allowbreak{}small & regression & 6,553 & 1,639 & 12 & 0 & \checkmark \\
credit & binary & 13,371 & 3,343 & 10 & 0 & \checkmark \\
Credit\_\allowbreak{}c & multiclass & 80,000 & 20,000 & 17 & 5 & \checkmark \\
credit\_\allowbreak{}reg & regression & 13,371 & 3,343 & 10 & 0 & \checkmark \\
Customer\_\allowbreak{}Personality\_\allowbreak{}Analysis & binary & 1,792 & 448 & 22 & 2 & \checkmark \\
customer\_\allowbreak{}satisfaction\_\allowbreak{}in\_\allowbreak{}airline & binary & 103,904 & 25,976 & 4 & 17 & \checkmark \\
dabetes\_\allowbreak{}130-us\_\allowbreak{}hospitals & binary & 81,412 & 20,354 & 8 & 12 &  \\
Data\_\allowbreak{}Science\_\allowbreak{}for\_\allowbreak{}Good\_\allowbreak{}Kiva\_\allowbreak{}Crowdfunding & multiclass & 536,964 & 134,241 & 7 & 4 & \checkmark \\
Data\_\allowbreak{}Science\_\allowbreak{}Salaries & regression & 3,004 & 751 & 1 & 4 &  \\
dataset\_\allowbreak{}sales & regression & 8,590 & 2,148 & 10 & 0 & \checkmark \\
debutanizer & regression & 1,915 & 479 & 7 & 0 & \checkmark \\
default\_\allowbreak{}of\_\allowbreak{}credit\_\allowbreak{}card\_\allowbreak{}clients & binary & 24,000 & 6,000 & 14 & 9 & \checkmark \\
delta\_\allowbreak{}ailerons & binary & 5,703 & 1,426 & 5 & 0 & \checkmark \\
delta\_\allowbreak{}elevators & regression & 7,613 & 1,904 & 6 & 0 & \checkmark \\
Diabetic\_\allowbreak{}Retinopathy\_\allowbreak{}Debrecen & binary & 920 & 231 & 16 & 3 & \checkmark \\
Diamonds & regression & 43,152 & 10,788 & 6 & 3 & \checkmark \\
drug\_\allowbreak{}consumption & multiclass & 1,507 & 377 & 12 & 0 &  \\
dry\_\allowbreak{}bean\_\allowbreak{}dataset & multiclass & 10,888 & 2,723 & 16 & 0 & \checkmark \\
E-CommereShippingData & binary & 8,799 & 2,200 & 6 & 4 & \checkmark \\
eeg-eye-state & binary & 11,984 & 2,996 & 14 & 0 & \checkmark \\
electricity & binary & 36,249 & 9,063 & 7 & 1 & \checkmark \\
elevators & regression & 13,279 & 3,320 & 18 & 0 & \checkmark \\
Employee & binary & 3,722 & 931 & 3 & 5 & \checkmark \\
estimation\_\allowbreak{}of\_\allowbreak{}obesity\_\allowbreak{}levels & multiclass & 1,688 & 423 & 8 & 8 & \checkmark \\
eye\_\allowbreak{}movements\_\allowbreak{}bin & binary & 6,086 & 1,522 & 20 & 0 & \checkmark \\
FICO-HELOC-cleaned & binary & 7,896 & 1,975 & 21 & 2 & \checkmark \\
fifa & regression & 14,450 & 3,613 & 5 & 0 & \checkmark \\
Firm-Teacher\_\allowbreak{}Clave-Direction\_\allowbreak{}Classification & multiclass & 8,640 & 2,160 & 0 & 16 & \checkmark \\
Fitness\_\allowbreak{}Club\_\allowbreak{}c & binary & 1,200 & 300 & 3 & 3 & \checkmark \\
Food\_\allowbreak{}Delivery\_\allowbreak{}Time & regression & 36,474 & 9,119 & 6 & 2 & \checkmark \\
FOREX\_\allowbreak{}audcad-day-High & binary & 1,467 & 367 & 10 & 0 & \checkmark \\
FOREX\_\allowbreak{}audcad-hour-High & binary & 35,060 & 8,765 & 10 & 0 & \checkmark \\
FOREX\_\allowbreak{}audchf-day-High & binary & 1,466 & 367 & 10 & 0 & \checkmark \\
FOREX\_\allowbreak{}audjpy-day-High & binary & 1,465 & 367 & 10 & 0 & \checkmark \\
FOREX\_\allowbreak{}audjpy-hour-High & binary & 35,060 & 8,765 & 10 & 0 & \checkmark \\
FOREX\_\allowbreak{}audsgd-hour-High & binary & 35,060 & 8,765 & 10 & 0 & \checkmark \\
FOREX\_\allowbreak{}audusd-hour-High & binary & 35,060 & 8,765 & 10 & 0 & \checkmark \\
FOREX\_\allowbreak{}cadjpy-day-High & binary & 1,467 & 367 & 10 & 0 & \checkmark \\
FOREX\_\allowbreak{}cadjpy-hour-High & binary & 35,060 & 8,765 & 10 & 0 & \checkmark \\
fried & regression & 32,614 & 8,154 & 10 & 0 & \checkmark \\
GAMETES\_\allowbreak{}Epistasis\_\allowbreak{}2-Way\_\allowbreak{}20atts\_\allowbreak{}0.1H\_\allowbreak{}EDM-1\_\allowbreak{}1 & binary & 1,280 & 320 & 0 & 20 &  \\
GAMETES\_\allowbreak{}Heterogeneity\_\allowbreak{}20atts\_\allowbreak{}1600\_\allowbreak{}Het\_\allowbreak{}0.4\_\allowbreak{}... & binary & 1,280 & 320 & 0 & 20 & \checkmark \\
garments\_\allowbreak{}worker\_\allowbreak{}productivity & regression & 957 & 240 & 8 & 5 & \checkmark \\
gas\_\allowbreak{}turbine\_\allowbreak{}CO\_\allowbreak{}and\_\allowbreak{}NOx\_\allowbreak{}emission & regression & 29,386 & 7,347 & 10 & 0 & \checkmark \\
Gender\_\allowbreak{}Gap\_\allowbreak{}in\_\allowbreak{}Spanish\_\allowbreak{}WP & multiclass & 3,796 & 950 & 13 & 0 &  \\
golf\_\allowbreak{}play\_\allowbreak{}dataset\_\allowbreak{}extended & binary & 876 & 219 & 3 & 6 & \checkmark \\
Goodreads-Computer-Books & regression & 987 & 247 & 5 & 0 &  \\
healthcare\_\allowbreak{}insurance\_\allowbreak{}expenses & regression & 1,070 & 268 & 3 & 3 & \checkmark \\
Heart-Disease-Dataset-(Comprehensive) & regression & 952 & 238 & 11 & 0 & \checkmark \\
heloc & binary & 8,000 & 2,000 & 22 & 0 & \checkmark \\
house\_\allowbreak{}16H & binary & 10,790 & 2,698 & 16 & 0 & \checkmark \\
house\_\allowbreak{}16H\_\allowbreak{}reg & regression & 18,227 & 4,557 & 16 & 0 & \checkmark \\
house\_\allowbreak{}8L & regression & 18,227 & 4,557 & 8 & 0 & \checkmark \\
house\_\allowbreak{}sales\_\allowbreak{}reduced & regression & 17,290 & 4,323 & 18 & 0 & \checkmark \\
houses & regression & 16,512 & 4,128 & 8 & 0 & \checkmark \\
housing\_\allowbreak{}price\_\allowbreak{}prediction & regression & 436 & 109 & 5 & 7 & \checkmark \\
HR\_\allowbreak{}Analytics\_\allowbreak{}Job\_\allowbreak{}Change\_\allowbreak{}of\_\allowbreak{}Data\_\allowbreak{}Scientists & binary & 15,326 & 3,832 & 3 & 10 & \checkmark \\
htru & binary & 14,318 & 3,580 & 8 & 0 & \checkmark \\
in\_\allowbreak{}vehicle\_\allowbreak{}coupon\_\allowbreak{}recommendation & binary & 10,147 & 2,537 & 0 & 21 & \checkmark \\
INNHotelsGroup & binary & 29,020 & 7,255 & 11 & 6 & \checkmark \\
Insurance & regression & 18,838 & 4,710 & 3 & 7 &  \\
internet\_\allowbreak{}firewall & multiclass & 52,425 & 13,107 & 7 & 0 & \checkmark \\
Intersectional-Bias-Assessment & regression & 8,800 & 2,200 & 14 & 5 & \checkmark \\
Is-this-a-good-customer & regression & 1,378 & 345 & 9 & 4 &  \\
JapaneseVowels & multiclass & 7,968 & 1,993 & 14 & 0 & \checkmark \\
jm1 & binary & 8,708 & 2,177 & 21 & 0 & \checkmark \\
jungle\_\allowbreak{}chess\_\allowbreak{}2pcs\_\allowbreak{}raw\_\allowbreak{}endgame\_\allowbreak{}complete & multiclass & 35,855 & 8,964 & 6 & 0 & \checkmark \\
Kaggle\_\allowbreak{}bike\_\allowbreak{}sharing\_\allowbreak{}demand\_\allowbreak{}challange & regression & 8,708 & 2,178 & 3 & 6 & \checkmark \\
kc1 & binary & 1,687 & 422 & 21 & 0 & \checkmark \\
kdd\_\allowbreak{}ipums\_\allowbreak{}la\_\allowbreak{}97-small & binary & 4,150 & 1,038 & 20 & 0 & \checkmark \\
kin8nm & regression & 6,553 & 1,639 & 8 & 0 & \checkmark \\
kr-vs-k & multiclass & 22,444 & 5,612 & 3 & 3 & \checkmark \\
kropt & multiclass & 22,444 & 5,612 & 0 & 6 & \checkmark \\
Laptop\_\allowbreak{}Prices\_\allowbreak{}Dataset & regression & 3,552 & 889 & 8 & 0 & \checkmark \\
law-school-admission-bianry & regression & 16,640 & 4,160 & 7 & 4 & \checkmark \\
led24 & multiclass & 2,560 & 640 & 0 & 24 & \checkmark \\
led7 & multiclass & 2,560 & 640 & 0 & 7 & \checkmark \\
letter & multiclass & 16,000 & 4,000 & 15 & 0 & \checkmark \\
Long & regression & 3,581 & 896 & 19 & 0 & \checkmark \\
MagicTelescope & binary & 15,216 & 3,804 & 9 & 0 & \checkmark \\
mammography & binary & 8,946 & 2,237 & 6 & 0 & \checkmark \\
maternal\_\allowbreak{}health\_\allowbreak{}risk & multiclass & 811 & 203 & 6 & 0 & \checkmark \\
mauna-loa-atmospheric-co2 & regression & 1,780 & 445 & 5 & 1 & \checkmark \\
mfeat-morphological & multiclass & 1,600 & 400 & 6 & 0 & \checkmark \\
MiamiHousing2016 & regression & 11,145 & 2,787 & 16 & 0 & \checkmark \\
microaggregation2 & multiclass & 16,000 & 4,000 & 20 & 0 & \checkmark \\
mobile\_\allowbreak{}c36\_\allowbreak{}oversampling & binary & 41,408 & 10,352 & 6 & 0 & \checkmark \\
Mobile\_\allowbreak{}Phone\_\allowbreak{}Market\_\allowbreak{}in\_\allowbreak{}Ghana & regression & 2,880 & 720 & 4 & 10 & \checkmark \\
Mobile\_\allowbreak{}Price\_\allowbreak{}Classification & multiclass & 1,600 & 400 & 14 & 6 & \checkmark \\
mozilla4 & binary & 12,436 & 3,109 & 4 & 0 & \checkmark \\
mv & regression & 32,614 & 8,154 & 7 & 3 & \checkmark \\
NASA\_\allowbreak{}PHM2008 & regression & 36,734 & 9,184 & 21 & 0 & \checkmark \\
national-longitudinal-survey-binary & binary & 3,926 & 982 & 9 & 7 & \checkmark \\
National\_\allowbreak{}Health\_\allowbreak{}and\_\allowbreak{}Nutrition\_\allowbreak{}Health\_\allowbreak{}Survey & binary & 1,822 & 456 & 7 & 0 & \checkmark \\
NHANES\_\allowbreak{}age\_\allowbreak{}prediction & regression & 1,821 & 456 & 4 & 3 & \checkmark \\
okcupid\_\allowbreak{}stem & multiclass & 21,341 & 5,336 & 2 & 11 & \checkmark \\
online\_\allowbreak{}shoppers & binary & 9,864 & 2,466 & 5 & 9 & \checkmark \\
page-blocks & multiclass & 4,378 & 1,095 & 10 & 0 & \checkmark \\
Parkinsons\_\allowbreak{}Telemonitoring & regression & 4,700 & 1,175 & 18 & 1 & \checkmark \\
pc1 & binary & 887 & 222 & 21 & 0 & \checkmark \\
pendigits & multiclass & 8,793 & 2,199 & 16 & 0 & \checkmark \\
Performance-Prediction & regression & 1,072 & 268 & 19 & 0 &  \\
phoneme & binary & 4,323 & 1,081 & 5 & 0 & \checkmark \\
Physicochemical\_\allowbreak{}r & regression & 36,584 & 9,146 & 9 & 0 & \checkmark \\
Pima\_\allowbreak{}Indians\_\allowbreak{}Diabetes\_\allowbreak{}Database & binary & 614 & 154 & 8 & 0 & \checkmark \\
puma8NH & regression & 6,553 & 1,639 & 8 & 0 & \checkmark \\
Pumpkin\_\allowbreak{}Seeds & binary & 2,000 & 500 & 12 & 0 & \checkmark \\
qsar\_\allowbreak{}aquatic\_\allowbreak{}toxicity & regression & 436 & 110 & 8 & 0 & \checkmark \\
qsar\_\allowbreak{}fish\_\allowbreak{}toxicity & regression & 726 & 182 & 4 & 2 & \checkmark \\
Rain\_\allowbreak{}in\_\allowbreak{}Australia & multiclass & 116,368 & 29,092 & 12 & 6 & \checkmark \\
rice\_\allowbreak{}cammeo\_\allowbreak{}and\_\allowbreak{}osmancik & binary & 3,048 & 762 & 7 & 0 & \checkmark \\
ringnorm & binary & 5,920 & 1,480 & 20 & 0 & \checkmark \\
rl & binary & 3,976 & 994 & 5 & 7 & \checkmark \\
SDSS17 & multiclass & 80,000 & 20,000 & 8 & 4 & \checkmark \\
segment & multiclass & 1,848 & 462 & 17 & 0 & \checkmark \\
seismic+bumps & binary & 2,067 & 517 & 14 & 4 &  \\
sensory & regression & 460 & 116 & 0 & 11 &  \\
shill-bidding & regression & 5,056 & 1,265 & 3 & 0 &  \\
Shipping & regression & 8,799 & 2,200 & 5 & 4 &  \\
Shop\_\allowbreak{}Customer\_\allowbreak{}Data & regression & 1,600 & 400 & 4 & 2 &  \\
shrutime & regression & 8,000 & 2,000 & 4 & 6 & \checkmark \\
shuttle & multiclass & 46,400 & 11,600 & 9 & 0 & \checkmark \\
Smoking\_\allowbreak{}and\_\allowbreak{}Drinking\_\allowbreak{}Dataset\_\allowbreak{}with\_\allowbreak{}body\_\allowbreak{}signal & binary & 793,076 & 198,270 & 22 & 1 & \checkmark \\
socmob & regression & 924 & 232 & 1 & 4 & \checkmark \\
space\_\allowbreak{}ga & regression & 2,485 & 622 & 6 & 0 & \checkmark \\
statlog & binary & 800 & 200 & 7 & 13 & \checkmark \\
steel\_\allowbreak{}industry\_\allowbreak{}energy\_\allowbreak{}consumption & regression & 28,032 & 7,008 & 7 & 3 & \checkmark \\
stock & regression & 760 & 190 & 9 & 0 & \checkmark \\
stock\_\allowbreak{}fardamento02 & regression & 5,021 & 1,256 & 5 & 1 &  \\
sulfur & regression & 8,064 & 2,017 & 6 & 0 & \checkmark \\
svmguide3 & regression & 994 & 249 & 22 & 0 & \checkmark \\
sylvine & binary & 4,099 & 1,025 & 20 & 0 & \checkmark \\
telco-customer-churn & binary & 5,634 & 1,409 & 3 & 15 & \checkmark \\
Telecom\_\allowbreak{}Churn\_\allowbreak{}Dataset & binary & 2,666 & 667 & 14 & 3 & \checkmark \\
thyroid & multiclass & 5,760 & 1,440 & 21 & 0 & \checkmark \\
thyroid-ann & multiclass & 3,017 & 755 & 21 & 0 & \checkmark \\
treasury & regression & 839 & 210 & 15 & 0 & \checkmark \\
twonorm & binary & 5,920 & 1,480 & 20 & 0 & \checkmark \\
vehicle & multiclass & 676 & 170 & 18 & 0 & \checkmark \\
VulNoneVul & regression & 4,553 & 1,139 & 16 & 0 &  \\
walking-activity & multiclass & 119,465 & 29,867 & 4 & 0 & \checkmark \\
wall-robot-navigation & multiclass & 4,364 & 1,092 & 24 & 0 & \checkmark \\
water\_\allowbreak{}quality & binary & 6,396 & 1,600 & 20 & 0 & \checkmark \\
Water\_\allowbreak{}Quality\_\allowbreak{}and\_\allowbreak{}Potability & binary & 2,620 & 656 & 8 & 0 &  \\
Waterstress & regression & 950 & 238 & 22 & 0 &  \\
waveform\_\allowbreak{}database\_\allowbreak{}generator & multiclass & 3,999 & 1,000 & 21 & 0 &  \\
waveform\_\allowbreak{}database\_\allowbreak{}generator\_\allowbreak{}version\_\allowbreak{}1 & multiclass & 4,000 & 1,000 & 21 & 0 & \checkmark \\
weather\_\allowbreak{}izmir & regression & 1,168 & 293 & 9 & 0 & \checkmark \\
website\_\allowbreak{}phishing & multiclass & 1,082 & 271 & 0 & 9 & \checkmark \\
Wilt & binary & 3,856 & 965 & 5 & 0 & \checkmark \\
wind & regression & 5,259 & 1,315 & 14 & 0 & \checkmark \\
wine & binary & 2,043 & 511 & 4 & 0 & \checkmark \\
wine+quality & regression & 5,197 & 1,300 & 11 & 0 & \checkmark \\
wine-quality-red & multiclass & 1,279 & 320 & 4 & 0 & \checkmark \\
wine-quality-white & multiclass & 3,918 & 980 & 11 & 0 & \checkmark \\
Wine\_\allowbreak{}Quality\_\allowbreak{}red & regression & 1,279 & 320 & 11 & 0 & \checkmark \\
Wine\_\allowbreak{}Quality\_\allowbreak{}white & regression & 3,918 & 980 & 11 & 0 & \checkmark \\
yeast & multiclass & 1,187 & 297 & 8 & 0 & \checkmark \\
\end{longtable}
\endgroup

\clearpage
\section{Resource Scaling and Compute}
\label{app:resources}

The real-world viability of a synthetic data algorithm hinges heavily on its resource consumption alongside its statistical utility, so we document the measured wall-clock time and memory usage for each method using a standardized protocol. All benchmarks process a single task per GPU at $\eps = 1$, timing the fitting and sampling phases together, with peak resident memory recorded from the operating system and peak GPU memory tracked via NVML.

\paragraph{Cross-method comparison.}
On benchmark datasets ranging from 760 to 63,000 training rows, \methodname{}'s cost is nearly flat: fit plus sampling takes 20 to 60 seconds, about 2 GB of resident memory, and under 1 GB of GPU memory. The reason is simple. The release is a single $O(n)$ pass over the data, and the fit only ever touches the $Lk$ released moments. On this range only PrivBayes (2 to 12 seconds, CPU only) and GEM are faster. Private-GSD goes from 52 seconds to a timeout at 63,000 rows, and 3-way AIM spikes on wide datasets. All numbers are averages over three seeds.

\begin{figure}[h]
  \centering
  \includegraphics[width=\linewidth]{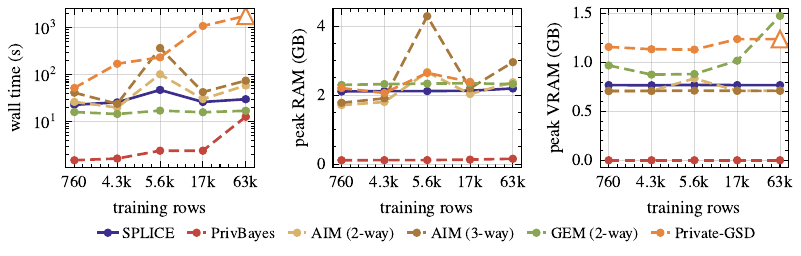}
  \caption{
Cost of generation at $\eps = 1$: total wall time (log scale; open triangles are timeouts), peak resident memory, and peak GPU memory. Datasets, left to right: stock, phoneme, telco-customer-churn, California-Housing-Classification, BNG(mv). Markers are means over three seeds with $\pm 1$ SEM error bars; a timed-out cell is a single run stopped at the limit.
  }
  \label{fig:resources}
\end{figure}

\paragraph{Per-method scaling laws.}
We repeat the measurement on a controlled grid built from one large mixed dataset: a row sweep from 12,500 to 400,000 rows at $d = 12$, and a column sweep from 6 to 24 columns at $n = 50,000$. Fitting $t \propto n^{\alpha} c^{d}$, where $\alpha$ is the row exponent and $c$ the per-column multiplier, the \methodname{} variants are essentially flat in both ($\alpha \leq 0.09$, $c \leq 1.05$) and finish in 30 to 120 seconds everywhere on the grid. GEM is similarly flat because it runs a fixed number of epochs. 2-way AIM grows moderately with $n$ ($\alpha \approx 0.55$); 3-way AIM does not depend on $n$ but pays a factor of about $1.16$ per column, going from 61 seconds at 6 columns to 960 seconds at 24. PrivBayes depends on both through its parent-set search, and scales as $n^{2.2}$ on wide tables. Private-GSD is the most expensive by far: 659 to 1,470 seconds across seeds at only 12,500 rows, the 2,400-second limit on two of three seeds at 25,000 rows, and a timeout beyond that even at 6 columns. So, among the methods we test, only the \methodname{} variants are both accurate on continuous features and insensitive to growth in either $n$ or $d$.

\begin{figure}[h]
  \centering
  \includegraphics[width=\linewidth]{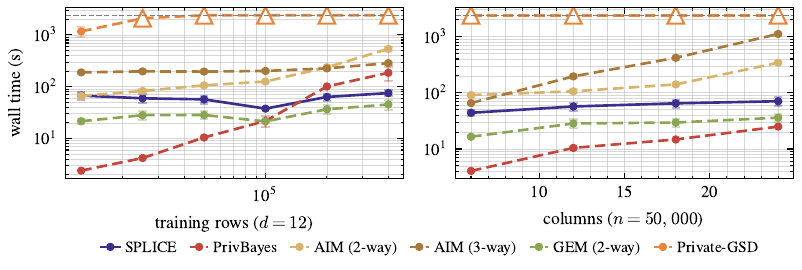}
  \caption{
Controlled scaling at $\eps = 1$: wall time against number of training rows at $d = 12$ (left, log-log) and against number of columns at $n = 50,000$ (right). Markers are means over three seeds with $\pm 1$ SEM error bars; open triangles are runs that hit the time limit.
  }
  \label{fig:scaling}
\end{figure}

\paragraph{Compute.}
All experiments ran on a single server with two AMD EPYC 9354 processors (64 cores, 128 threads), 512 GB of RAM, and six NVIDIA RTX PRO 6000 Blackwell GPUs with 96 GB of memory each.
Every generation task fits comfortably on one GPU; the resource measurements of \Cref{app:resources} run one task at a time per GPU, while the sweeps schedule independent tasks across GPUs in parallel.
\methodname{} itself is light: a single fit-plus-sample cell takes 20 to 120 seconds and under 1 GB of GPU memory, so its full 16,800-cell sweep totals roughly 200 GPU-hours and would run on one consumer-grade GPU in about a week.
The overall benchmark cost is dominated by the baselines.

\end{document}